\documentclass[10pt,twocolumn,letterpaper]{article}

\usepackage{iccv}
\usepackage{times}
\usepackage{epsfig}
\usepackage{graphicx}
\usepackage{amsmath}
\usepackage{amssymb}

\usepackage[accsupp]{axessibility}  %
\usepackage{mypackages}
\def\datasetname{TREK-150} 
\def\datasetlink{\href{https://machinelearning.uniud.it/datasets/trek150/}{\footnotesize\texttt{https://machinelearning.uniud.it/datasets/trek150/}}}
\def\videolink{\href{https://youtu.be/oX1nICHgEJM}{\footnotesize\texttt{https://youtu.be/oX1nICHgEJM}}}

\newcommand{\AF}[1]{\textcolor{black}{#1}}
\newcommand{\pgraph}[1]{\paragraph{#1}}

\usepackage[pagebackref=true,breaklinks=true,letterpaper=true,colorlinks,bookmarks=false]{hyperref}

\iccvfinalcopy %

\def\iccvPaperID{15} %

\ificcvfinal\fi

\begin{document}

\title{Is First Person Vision Challenging for Object Tracking?}

\author{Matteo Dunnhofer$^{\bullet}$\\
\and
Antonino Furnari$^{\star}$
\and
Giovanni Maria Farinella$^{\star}$
\and
Christian Micheloni$^{\bullet}$ \and
$^{\bullet}$Machine Learning and Perception Lab, University of Udine, Udine, Italy \\
$^{\star}$Image Processing Laboratory, University of Catania, Catania, Italy
}

\maketitle
\ificcvfinal\thispagestyle{empty}\fi

\begin{abstract}
   Understanding 
human-object interactions is 
fundamental 
in First Person Vision (FPV). 
Tracking algorithms \AF{which}
follow the objects \AF{manipulated by}
the camera wearer 
\AF{can provide useful cues to effectively model such interactions.}
Visual tracking solutions available in the computer vision literature have 
\AF{significantly improved their performance}
in the last years for a large variety of target objects and tracking scenarios.
However, despite 
\AF{a few previous attempts to exploit trackers in FPV applications, a methodical analysis of the performance of state-of-the-art trackers in this domain is still missing.}
In this paper, we fill the gap by presenting the first systematic study of object tracking in FPV. Our study extensively analyses the performance of recent visual trackers and baseline FPV trackers with respect to different aspects and considering a new performance measure. This is achieved through \datasetname, a novel benchmark dataset composed of 150 densely annotated video sequences.
Our results show that object tracking in FPV is challenging\AF{, which suggests that more research efforts should be devoted to this problem so that tracking could benefit FPV tasks.}

\end{abstract}

\section{Introduction}
Understanding the interactions between a camera wearer and the surrounding objects is a fundamental problem in First Person Vision (FPV)~\AF{\cite{EK55,Wang2020,liu2020forecasting,RULSTMpami,damen2016you,ragusa2020meccano,cai2016understanding,gberta_2017_RSS,bertasius2017unsupervised,Rhoi2020}}.
\AF{To model such interactions,}
the continuous knowledge of where \AF{an object of interest} is located inside the video frame \AF{is advantageous}.
\AF{The benefits of tracking in FPV have been explored by a few previous}
works  %
\AF{to predict future active objects~\cite{Furnari2017}, analyze social interactions~\cite{Aghaei2016icpr}, improve the performance of hand detection for rehabilitation purposes~\cite{Visee2020}, locate hands and capture their movements for action recognition~\cite{kapidis2019egocentric} and human-object interaction forecasting~\cite{liu2020forecasting}.}
On a more abstract level, the features computed after the frame by frame localization of objects have been increasingly used for egocentric action recognition \cite{Wang2020,wu2019long,ma2016going} and anticipation \cite{RULSTMpami,rodin2021predicting,sener2020temporal}.

\begin{figure}[t]%
\centering
\includegraphics[width=\columnwidth]{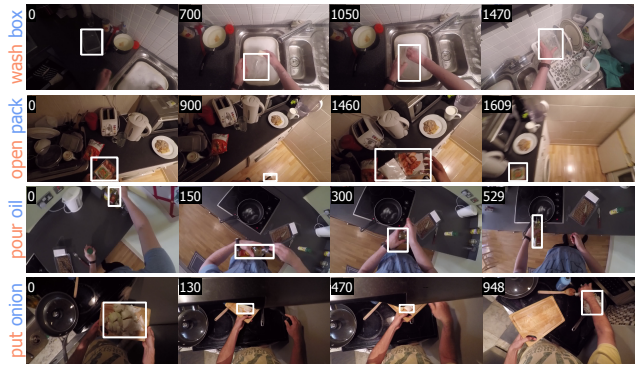}
\caption{Qualitative examples of some sequences contained in the proposed \datasetname\ benchmark dataset. The white rectangle represents the ground-truth bounding box \AF{of} the target object. Each number in the top left corner identifies the frame index. For each sequence, the action performed by the camera wearer is also reported (verb in orange, noun in blue). \AF{As can be noted, objects undergo significant appearance and state changes due to the manipulation by the camera wearer, which makes the proposed setting challenging for current trackers.}}
\label{fig:examples}
\vspace{-0.5em}
\end{figure}

\AF{Despite the aforementioned attempts to leverage tracking in egocentric vision pipelines, most approaches rely on}
object detection models that evaluate video frames 
\AF{independently.}
This \AF{paradigm}
has the drawback of ignoring all the temporal information coming from the object appearance and motion contained in consecutive video frames \AF{and generally requires a higher computational cost due to the repeated detection process on every frame.}
In contrast, visual object tracking aims to exploit past information about the target \AF{to}
infer its position and shape in the next frames of a video~\AF{\cite{Maggio2011}}. 
This process is subject to different challenges 
\AF{including}
occlusions, appearance changes, illumination variation, fast motion, and motion blur. Additionally, many practical applications
pose real-time constraints to the computation, \AF{which specifically hold in FPV when the localization of objects is needed by higher-level real-time algorithms.}
\AF{While the use cases of object tracking in egocentric vision are manifold as previously discussed, it is clear that tracking is still not a dominant technology in the FPV field. We experimentally show that this is mainly due to the limited performance of current trackers in egocentric videos due to the involved FPV challenges such as camera motion, persistent occlusion, significant scale and state changes, as well as motion blur (see Figure~\ref{fig:examples}).}
\AF{Due to these challenges, previous works have proposed customized approaches to}
track 
specific targets like people~\cite{Alletto2015}, people faces~\cite{Aghaei2016}, or hands~\cite{kapidis2019egocentric,Visee2020,Mueller2017,Han2020,Sun2010} \AF{from the FPV perspective}. 
\AF{A solution specifically designed to track arbitrary objects in egocentric videos is still missing.}
Instead, the computer vision community has 
\AF{made significant}
progress in the visual tracking of generic objects. 
This 
\AF{has been possible}
thanks to development of new and effective \AF{tracking} principles \cite{MOSSE,KCF,SiamFC,ECO,DiMP,SiamBAN,Ocean,LTMU}, and to the careful design of benchmark datasets~\cite{OTB,UAV123,NfS,NUSPRO,LaSOT,GOT10k} and challenges \cite{VOT2015,VOT2017,VOT2019,VOT2020}. Nowadays, the state-of-the-art tracking solutions achieve excellent results on a large variety of tracking domains \cite{OTB,UAV123,NfS,GOT10k,VOT2019,VOT2020}.
However, \AF{all these research endeavours have taken into account mainly the classic third person scenario in which objects are observed from an external point of view and are not manipulated by the camera wearer. Additionally, the performance of existing trackers has never been evaluated in the FPV domain, which raises the question of whether current solutions can be used ``off-the-shelf'' or more domain-specific investigations should be carried out.}

\AF{To answer the aforementioned questions, in this paper we aim to}
extensively analyze the problem of visual object tracking in \AF{the} FPV \AF{domain}. 
\AF{Given the lack of suitable benchmarks,} we follow the standard practice of the visual tracking community that suggests to build an accurate dataset for evaluation \cite{OTB,TC128,UAV123,NUSPRO,NfS,VOT2019,CDTB}. Therefore, we propose a novel visual tracking benchmark, \datasetname\ (TRacking-Epic-Kitchens-150), which is obtained from the large and challenging FPV dataset EPIC-KITCHENS-55 (EK-55) \cite{EK55}. \datasetname\ provides 150 video sequences densely annotated with the bounding boxes of a %
target object the camera wearer interacts with. Additionally, sequences
\AF{have been}
labelled with attributes that identify the visual changes 
\AF{the object is undergoing, the class of the target object}
and the action the person is performing. 
Using the dataset, 
we present an in-depth study of the accuracy and speed performance of both non-FPV and FPV visual trackers.
A new performance measure is also introduced to evaluate trackers with respect to FPV scenarios.

In sum, the contributions of this paper are: (i) the first systematic analysis of visual object tracking in FPV; (ii) the description and release of the new \datasetname\ dataset, which offers new challenges and complementary features with respect to existing visual tracking benchmarks; (iii) two FPV baseline trackers combining a state-of-the-art generic object tracker and FPV object detectors; (iv) a new and improved measure to assess the tracker's ability to maintain temporal reference to targets.

Our results show that FPV offers challenging tracking scenarios for the most recent \AF{and} accurate trackers \cite{LTMU,ATOM,VITAL,ECO,KYS} and even for FPV trackers. \AF{Considering the potential impact of tracking on FPV, we suggest that more research efforts should be devoted to the considered task, for which we believe the proposed \datasetname\ benchmark will be a key research tool.}
Annotations, trackers' results, and code are available at \datasetlink.

\begin{table*}[t]
\fontsize{8}{9}\selectfont
	\centering
	\caption{Statistics of the proposed \datasetname\ benchmark compared with other benchmarks designed for SOT evaluation.}
	\label{tab:datasets}
	\begin{tabular}{l | c  c  c  c  c  c  c  c | c }
		\toprule
		\multirow{2}{*}{Benchmark} & OTB-50 & OTB-100 & TC-128 & UAV123  & NUS-PRO & NfS  & VOT2019 & CDTB  & \multirow{2}{*}{\datasetname}\\
                    & \cite{OTB2013} & \cite{OTB} & \cite{TC128} & \cite{UAV123}  & \cite{NUSPRO} & \cite{NfS}  & \cite{VOT2019} & \cite{CDTB}  \\
		\midrule
		\# videos & 51 & 100 & 128 & 123 & 365 & 100 & 60 & 80 & 150 \\
		\# frames & 29K & 59K & 55K & 113K & 135K & 383K & 20K & 102K & 97K \\
		Min frames \AF{across videos} & 71 & 71 & 71 & 109 & 146 & 169 & 41 & 406 & 161  \\
		Mean frames \AF{across videos} & 578 & 590 & 429 & 915 & 371 & 3830 & 332 & 1274 & 649 \\
		Median frames \AF{across videos} & 392 & 393 & 365 & 882 & 300 & 2448 & 258 & 1179 & 484 \\
		Max frames \AF{across videos} & 3872 & 3872 & 3872 & 3085 & 5040 & 20665 & 1500 & 2501 & 4640 \\
		Frame rate & 30 FPS & 30 FPS & 30 FPS  & 30 FPS  & 30 FPS  & 240 FPS  & 30 FPS  & 30 FPS  & 60 FPS  \\
		\# target object classes & 10 & 16 & 27  & 9  & 8 & 17  & 30  & 23  & 34  \\
		\# sequence attributes & 11 & 11 & 11 & 12  & 12 & 9 & 6  & 13  & 17  \\
		FPV & \xmark & \xmark & \xmark & \xmark & \xmark & \xmark & \xmark & \xmark & \cmark \\
		\# action verbs & n/a & n/a & n/a & n/a & n/a & n/a & n/a & n/a & 20 \\
		\bottomrule		
\end{tabular}
\end{table*}

\section{Related Work}

\pgraph{Visual Tracking in FPV.}
\label{sec:fpvtrackers}
There have been some attempts to tackle visual tracking in FPV.
Alletto et al.~\cite{Alletto2015} improved the TLD tracker~\cite{TLD} with a 3D odometry based module to track people.
For a similar task, Nigam et al.~\cite{Nigam2017} proposed a combination of the Struck~\cite{Struck} and MEEM~\cite{MEEM} trackers with a person re-identification module.
Face tracking was tackled by Aghaei et al. \cite{Aghaei2016} through %
\AF{a}
multi-object tracking approach 
\AF{termed}
extended-bag-of-tracklets.
Hand tracking was studied in 
\AF{several}
works\AF{~\cite{kapidis2019egocentric,Visee2020,Mueller2017,Han2020,Sun2010}}. Sun et al. \cite{Sun2010} developed a particle filter framework for hand pose tracking. M{\"u}ller et al. \cite{Mueller2017} proposed a solution based on an RGB camera and a depth sensor. Kapidis et al. \cite{kapidis2019egocentric} and Vis\'ee et al. \cite{Visee2020} proposed to combine the YOLO \cite{YOLO} detector trained for hand detection with trackers. The
\AF{former}
used the multi-object tracker DeepSORT~\cite{DeepSORT}, whereas the 
\AF{latter}
employed the KCF \cite{KCF} single object tracker. Han et al. \cite{Han2020} exploited a detection-by-tracking approach on video frames acquired with 4 fisheye cameras.
All the presented solutions focused on tracking specific targets (i.e., people, faces, or hands), and thus they \AF{are likely to} fail in generalizing to arbitrary target objects. 
Moreover, they have been validated on custom designed datasets, \AF{which limits their reproducibility and the ability to compare them to other works}. 
\AF{In contrast,} we focus on the evaluation of algorithms \AF{for the generic object tracking task}.
\AF{We design our evaluation to be reproducible and extendable by releasing \datasetname, a dataset of 150 videos of different objects manipulated by the camera wearer, which we believe will be useful to study object tracking in FPV.}
To the best of our knowledge, 
\AF{ours is the first attempt}
to evaluate systematically generic object tracking in \AF{the FPV context}.

\pgraph{Visual Tracking for Generic Settings.}
In recent years, there has been an increased interest in developing accurate and robust single object tracking (SOT) algorithms for generic targets and domains.
Preliminary trackers were based on mean shift algorithms~\cite{Comanciu2000}, key-point~\cite{Matrioska}, part-based methods \cite{LGT,OGT}, or SVM learning~\cite{Struck}. Later, \AF{solutions based on} correlation filters gained popularity thanks to their processing speed~\cite{MOSSE,KCF,DSST,Staple,BACF}. More recently, \AF{algorithms based on} deep learning have been \AF{proposed} to extract efficient image and object features. This kind of representation has been used in deep regression networks~\cite{GOTURN,Dunnhofer2021ral}, online tracking-by-detection methods~\cite{MDNet,VITAL}, \AF{approaches} based on reinforcement learning~\cite{Yun2017,Chen2018,Dunnhofer2019,Guo2020tip}, deep discriminative correlation filters~\cite{ECO,ATOM,DiMP,PrDiMP,D3S,KYS}, and \AF{trackers based on} siamese networks~\cite{SiamFC,SiamRPNpp,SiamMask,SiamBAN,Ocean}. 
All these methods have been designed for tracking arbitrary target objects in unconstrained domains. However, no solution has been studied and validated on a number of diverse FPV sequences \AF{as} we propose in this paper.

\begin{figure*}[t]%
\centering
\includegraphics[width=\linewidth]{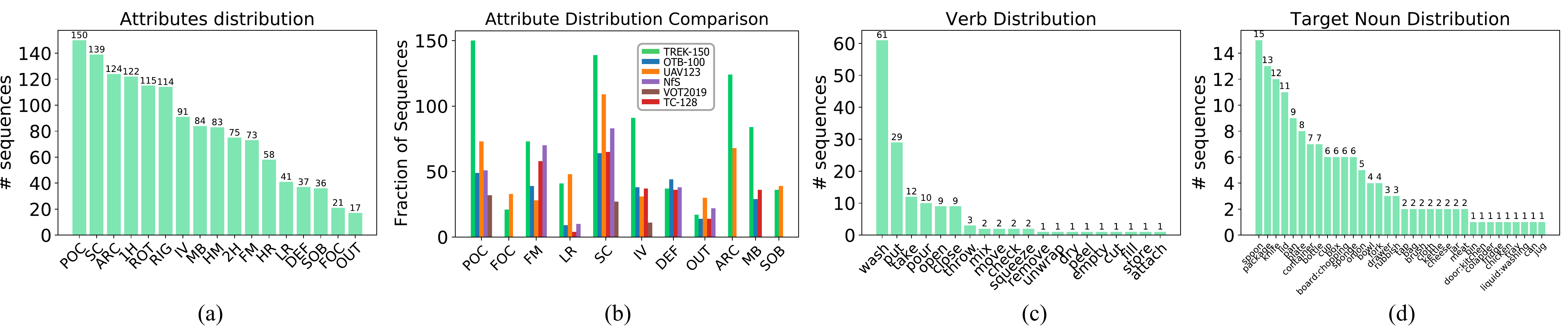}
\caption{(a) Distribution of the sequences within \datasetname\ with respect to the attributes. (b) Comparison of \AF{the} distributions \AF{of} common attributes \AF{in} different benchmarks. Distributions of (c) action verb labels, and (d) target object categories (nouns).}
\label{fig:distributions}
\vspace{-1.15em}
\end{figure*}

\pgraph{Visual Tracking Benchmarks.}
Disparate bounding box level benchmarks are available today to evaluate the performance of SOT algorithms. 
The Object Tracking Benchmarks (OTB) OTB-50 \cite{OTB2013} and OTB-100 \cite{OTB} are two of the most popular benchmarks in the visual tracking community. They provide 51 and 100 sequences respectively \AF{including} generic targets like vehicles, people, faces, toys, characters, etc.
The Temple-Color 128~(TC-128) dataset \cite{TC128} comprises 128 videos and was designed for the evaluation of color-enhanced trackers. 
The UAV123 dataset \cite{UAV123} was constructed to benchmark the tracking of 9 classes of target in 123 videos captured by unmanned aerial vehicle (UAV) cameras.
The NUS-PRO dataset~\cite{NUSPRO} contains 365 sequences and aims to benchmark human and rigid object tracking with targets belonging to one of 8 categories.
The Need for Speed (NfS) dataset~\cite{NfS} provides 100 sequences with a frame rate of 240 FPS. The aim of the authors was to benchmark the effects of frame rate variations on the tracking performance.
The VOT2019 benchmark~\cite{VOT2019} was the last iteration of the annual Visual Object Tracking challenge that required bounding-boxes as target object representation. This dataset contains 60 highly challenging videos, with generic target objects belonging to 30 different categories. 
The Color and Depth Tracking Benchmark (CDTB) dataset \cite{CDTB} offers 80 RGB sequences paired with a depth channel. This benchmark aims to explore the use of depth information to improve tracking performance. 
Following the increased development of deep learning based trackers, large-scale generic-domain SOT datasets have been recently released~\cite{TrackingNet,GOT10k,LaSOT}. These include more than \AF{a} thousand videos normally split into training and test subsets. The evaluation protocol \AF{associated with} these sets requires the evaluation of the trackers after \AF{they have been trained on}
the provided training set.
Despite the fact that all the presented benchmarks offer various tracking scenarios, 
\AF{limited work has focused on FPV, with some studies tackling the problem of tracking pedestrians or cars from a moving camera~\cite{sanchez2019predictor}.}
\AF{Some datasets of egocentric videos such as ADL~\cite{pirsiavash2012detecting} and EK-55~\cite{EK55} contain bounding-box object annotations. But due to the sparse nature of such annotations (typically 1/2 FPS), these datasets cannot be used for the accurate evaluation of trackers in FPV context.}
To the best of our knowledge, our proposed \datasetname\ dataset is the first benchmark for tracking objects \AF{which are relevant to (or manipulated by) a camera wearer} in egocentric videos. We believe that \datasetname\ is tantalizing for the tracking community because it offers complementary tracking situations (\AF{which we characterize}
with a total of $17$ attributes) and new target object categories (for a total of $34$) that are not present in other tracking benchmarks. 
Since in this paper we aim to benchmark generic approaches to visual tracking (that would not necessarily \AF{consider the} deep learning approach), we follow the practice of \AF{previous works} \cite{OTB,TC128,UAV123,NUSPRO,NfS,VOT2019,CDTB} and set up a well described dataset for evaluation of generic SOT algorithms. We believe that \datasetname\ can be a useful research tool for both the FPV and visual tracking research communities.

\section{The \datasetname\ Benchmark Dataset}

The proposed \datasetname\ dataset is composed of 150 video sequences.
In each, a single target object is labeled with a bounding box which encloses the visible parts of the object. The bounding boxes are given for each frame in which the object is visible (as a whole or in part). 
To be compliant with other tracking challenges, every sequence is additionally labeled with one or more of 17 attributes describing the visual variability of the target in the sequence, 
\AF{plus two additional action verb and noun attributes indicating the action performed by the camera wearer and the class of the target.}
Qualitative examples of the video sequences are shown in Figure \ref{fig:examples}, whereas Table \ref{tab:datasets} \AF{reports} key statistics of our dataset in comparison with \AF{existing} benchmarks.\footnote{Please see Appendix \ref{sec:datasetdiff} of the supplementary material for additional motivations and details.}

\begin{table}[t]
\fontsize{6}{7}\selectfont
	\centering
	\caption{Selected sequence attributes. The first block of rows describes attributes commonly used by the visual tracking community. The last four rows describe additional attributes \AF{introduced in this paper to characterize} FPV tracking sequences.}
	\label{tab:attrdesc}
	\setlength\tabcolsep{.15cm}
	\begin{tabular}{m{3.5em} | m{33em} }
		\toprule
		Attribute & Meaning \\
		\midrule
		SC & \underline{Scale Change}: the ratio of the bounding-box area of the first and the current frame is outside the range [0.5, 2] \\
		ARC & \underline{Aspect Ratio Change}: the ratio of the bounding-box aspect ratio of the first and the current frame is outside the range [0.5, 2] \\
		IV & \underline{Illumination Variation}: the area of the target bounding-box is subject to light variation \\
		SOB & \underline{Similar Objects}: there are objects in the video of the same object category or with similar appearance to the target \\
		RIG & \underline{Rigid Object}: the target is a rigid object \\
		DEF & \underline{Deformable Object}: the target is a deformable object \\
		ROT & \underline{Rotation}: the target rotates in the video \\
		POC & \underline{Partial Occlusion}: the target is partially occluded in the video \\
		FOC & \underline{Full Occlusion}: the target is fully occluded in the video \\
		OUT & \underline{Out Of View}: the target completely leaves the video frame \\
		MB & \underline{Motion Blur}: the target region is blurred due to target or camera motion \\
		FM & \underline{Fast Motion}: the target bounding-box has a motion change larger than its size \\
		LR & \underline{Low Resolution}: the area of the target bounding-box is less than 1000 pixels in at least one frame \\
		\midrule
		HR & \underline{High Resolution}: the area of the target bounding-box is larger than 250000 pixels in at least one frame \\
		HM & \underline{Head Motion}: the person moves their head significantly thus causing camera motion \\
		1H & \underline{1 Hand Interaction}: the person interacts with the target object with one hand for consecutive video frames \\
		2H & \underline{2 Hands Interaction}: the person interacts with the target object with both hands for consecutive video frames \\
		\bottomrule		
\end{tabular}
\end{table}

\pgraph{Data Collection.}
The videos have been sampled from EK-55 \cite{EK55}, \AF{which is a public, large-scale, and diverse dataset of egocentric videos focused on human-object interactions in kitchens.}
\AF{EK-55} provides videos annotated with the actions performed by the \AF{camera wearer} in the form of temporal bounds and verb-noun labels.
\AF{The dataset also contains sparse bounding-box references of manipulated objects annotated at 2 frames per second in a temporal window around each action.}
To obtain a suitable pool of video sequences interesting for object tracking, we cross-referenced the original verb-noun temporal annotations of EK-55 to the sparse bounding box labels. This allowed to select sequences in which the camera wearer manipulates an object.
Each sequence is composed of the video frames contained \AF{within the temporal bounds of} the action, extracted \AF{at} the original 60 FPS frame rate and at the original \AF{full HD} frame size~\cite{EK55}. According to the authors of~\cite{EK55}, this frame rate is necessary in FPV to contrast the fast motion and motion blur happening due to the proximity of the main scene and the camera point of view.
\AF{From the initial pool, we selected 150 video sequences which were characterized by attributes such as scale changes, partial/full occlusion and fast motion, which are commonly considered in standard tracking benchmarks~\cite{OTB,UAV123,TrackingNet,LaSOT,VOT2019}. 
The top part of Table~\ref{tab:attrdesc} reports the $13$ attributes considered for the selection.}
 
\pgraph{Data Labeling.}
\label{sec:labeling}
After selection, the 150 sequences \AF{were associated to only} %
3000 bounding boxes, due to the sparse \AF{nature of the} object annotations \AF{in} EK-55. 
Since it has been shown that visual tracking benchmarks require dense and accurate annotations~\cite{VOT2019,UAV123,LaSOT,OxUvA}, we re-annotated the bounding boxes of the target objects on the 150 sequences. 
Batches of sequences were delivered to annotators who were explicitly instructed to perform the labeling. 
Such initial annotations were then carefully checked and refined by a visual tracking expert. This process produced 97296 frames labeled with bounding boxes related to the position and visual presence of objects the camera wearer is interacting with.
Following the initial annotations, we employed axis-aligned bounding boxes. This kind of representation is widely used in many FPV pipelines \cite{Furnari2017,RULSTMiccv,RULSTMpami,EK55,Kapidis2019,Visee2020,Shan2020}, and thus it allows us to give immediate results on the impact of trackers in such contexts. Moreover, the recent progress of trackers on various benchmarks that use this state representation \cite{OTB,UAV123,NfS,CDTB,TrackingNet,LaSOT,GOT10k} demonstrates that it provides sufficient information about the target for consistent and reliable performance evaluation.

Along with the bounding boxes, the sequences have been labeled considering 17 attributes which define the \AF{motion and visual appearance changes} the target object is subject.
\AF{These include the aforementioned 13 standard tracking attributes, plus 4 additional ones (High Resolution, Head Motion, 1-Hand Interaction, 2-Hands Interaction) which have been introduced to characterize FPV sequences and are summarized in the bottom part of Table~\ref{tab:attrdesc}.}
\AF{Figure~\ref{fig:distributions}(a) reports the distributions of the sequences with respect to the 17 attributes.}
\AF{Figure~\ref{fig:distributions}(b) compares the distributions of the most common SOT attributes in} \datasetname\ and in other well-known benchmarks.
Our dataset provides a larger number of sequences \AF{affected by partial occlusions}~(POC), \AF{changes in scale}~(SC) and/or aspect ratio~(ARC), and motion blur~(MB). 
We claim that these peculiarities, which are complementary to those of existing datasets, are due to the particular first person viewpoint, camera motion, and the human-object interactions contained in the videos.
\AF{Based on EK-55's verb-noun labels,} sequences were also associated to 
\AF{20 verb labels~(e.g., ``wash'' - see Figure~\ref{fig:examples}) and 34 noun labels indicating the category of the target object (e.g., ``box'').}
\AF{Figures~\ref{fig:distributions}(c-d) show the distributions of the videos relative to 
verbs and target nouns. As can be noted, \datasetname\ reflects the EK-55's long-tail distribution of labels.}

\section{Trackers}
We considered
33 trackers in our benchmark evaluation. 
\AF{31 of these trackers}
have been selected 
\AF{to represent}
different \AF{popular} approaches to SOT, \AF{for instance with respect to the} %
matching strategy, type of image representations, learning strategy, etc.
\AF{Specifically, in the analysis we} have included short-term trackers \cite{VOT2019} 
based on both correlation-filters with hand-crafted features (MOSSE \cite{MOSSE}, DSST \cite{DSST}, KCF \cite{KCF}, Staple \cite{Staple}, BACF \cite{BACF}, DCFNet \cite{DCFNet}, STRCF \cite{STRCF}, MCCTH \cite{MCCTH}) 
\AF{and}
deep features (ECO \cite{ECO}, ATOM \cite{ATOM}, DiMP \cite{DiMP}, PrDiMP \cite{PrDiMP}, KYS \cite{KYS}). We also considered deep siamese networks (SiamFC \cite{SiamFC}, GOTURN \cite{GOTURN}, DSLT \cite{DSLT}, SiamRPN++ \cite{SiamRPNpp}, SiamDW \cite{SiamDW}, UpdateNet \cite{UpdateNet}, SiamFC++ \cite{SiamFCpp}, SiamBAN \cite{SiamBAN}, Ocean \cite{Ocean}), tracking-by-detection methods (MDNet \cite{MDNet}, VITAL \cite{VITAL}), as well as trackers based on target segmentation representations (SiamMask \cite{SiamMask}, D3S \cite{D3S}), meta-learning (MetaCrest \cite{MetaTrackers}), and fusion strategies (TRASFUST \cite{Dunnhofer2020accv}).
The long-term \cite{VOT2019} trackers SPLT \cite{SPLT}, GlobalTrack \cite{GlobalTrack}, and LTMU \cite{LTMU} have been also taken into account in the study. 
These trackers are designed to address longer target occlusion and out of view periods by exploiting object re-detection modules. 
All of the selected trackers 
\AF{are state-of-the-art approaches published}
between the years 2010-2020.

In addition to the aforementioned generic object trackers, we developed 2 baseline FPV trackers that combine the LTMU tracker \cite{LTMU} with (i) the EK-55 trained Faster-R-CNN \cite{EK55} and (ii) the Faster-R-CNN-based hand-object detector \cite{Shan2020}.
We refer to them as LTMU-F and LTMU-H respectively.
These baseline trackers exploit the respective detectors as object re-detection modules according to the LTMU scheme \cite{LTMU}. In short, the re-detection happens when a verification module notices that the tracker is not following the correct target. In such a case, the module triggers the execution of the respective FPV detector which proposes candidate locations of the target object. Each of the candidates is evaluated by the verification module, and the location with highest confidence is used to re-initialize the tracker.\footnote{More details are given in Appendix \ref{sec:trackerdetails} of the supplementary material.}
The two modules implement conceptually different strategies for FPV-based object localization. The first aims to find objects in the scene, while the second looks for the interaction between the camera wearer and objects.

\section{Evaluation}
\label{sec:exp}

\pgraph{Evaluation Protocols.}
We employed three standard protocols to perform our analysis.\footnote{See Appendix \ref{sec:expdetail} of the supplementary material for further details.}
The first is the one-pass evaluation (OPE) protocol detailed in \cite{OTB}, \AF{which}
implements the most realistic way to execute trackers. It consists in initializing a tracker with the ground-truth bounding box \AF{of}
the target in the first frame and let the tracker run \AF{on}
every 
\AF{subsequent}
frame until the end of the sequence.

To obtain a more robust evaluation \cite{Kristan2016}, especially for the analysis over sequence attributes and action verbs, we employ the recent protocol of \cite{VOT2020} which defines different points of initialization along a sequence. A tracker is initialized with the ground-truth in each point and let run either forward or backward in time (depending on the longest sub-sequence yielded by the initialization point) until the end of the sub-sequence.
This protocol allows a tracker to better cover all the situations happening in the sequences, ultimately leading to more robust evaluation scores.
We refer to this setup as multi-start evaluation (MSE).

Since many FPV tasks such as object interaction \cite{damen2016you} and early action recognition \cite{RULSTMiccv}, or action anticipation \cite{EK55}, require real-time computation, we evaluated the ability of trackers to provide their object localization in such a setting. This was achieved by following the details given in \cite{VOT2017,Li2020}. In short, this protocol, which we refer to as RTE, runs an algorithm considering its running time. The protocol skips all the frames, considered to occur regularly according to the frame rate, which appeared during the interval between the algorithm's execution start and end times.

\begin{figure*}[t]%
\centering
\includegraphics[width=\linewidth]{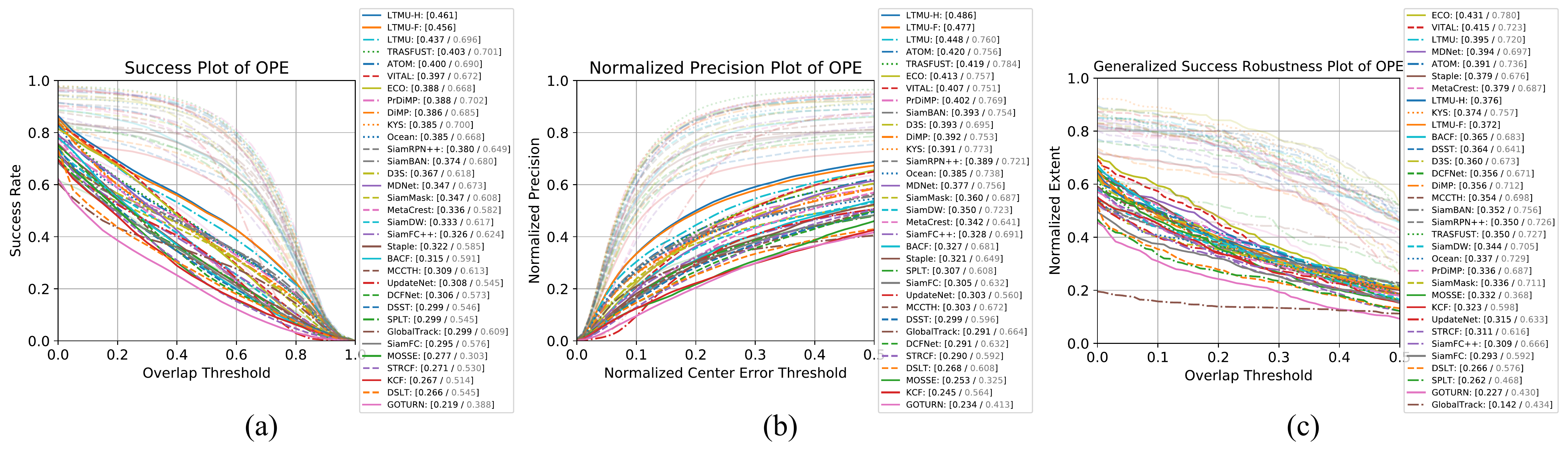}
\caption{Performance of the selected trackers on the proposed \datasetname\ benchmark under the OPE protocol. The curves in solid colors report the performance of the 33 benchmarked trackers on \datasetname,\ whereas the curves overlaid in semi-transparent colors outline the performance obtained by the same trackers on the standard OTB-100~\cite{OTB} dataset. \AF{In brackets, next to the trackers' names, we report the SS, NPS and GR values achieved on \datasetname\ (in black) and on OTB-100 \cite{OTB} (in gray)}.  
As can be noted, all the trackers \AF{exhibit} a significant performance drop when 
\AF{tested on our challenging FPV benchmark}.
LTMU-H and LTMU-F achieve marginally better performance, while we expect significant boosts to be achievable with a careful design of FPV trackers. }
\label{fig:results}
\end{figure*}

\begin{figure*}[t]%
\centering
\includegraphics[width=\linewidth]{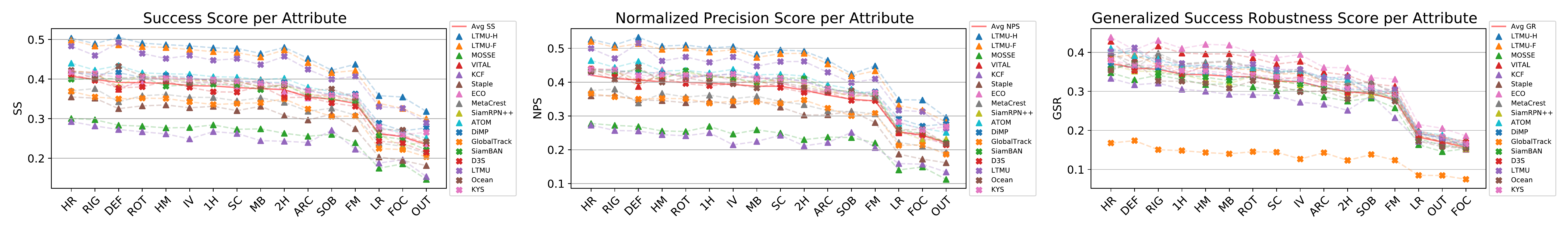}
\caption{SS, NPS, and GSR of 17 of the benchmarked trackers on the sequence attributes of proposed \datasetname\ benchmark under the MSE protocol. The red plain line highlights the average performance. (The results for POC are not reported because this attribute is present in every sequence).}
\label{fig:resultsattributes}
\end{figure*}

\pgraph{Performance Measures.}
To quantitatively assess the performance of the trackers on the proposed dataset, we used different measures that compare all tracker's predicted bounding boxes with respect to the temporally aligned ground-truth \AF{bounding boxes}. 
To evaluate the \AF{localization accuracy of the trackers, we employ}
the success plot \cite{OTB}, which shows the percentage of \AF{predicted} bounding boxes whose intersection-over-union with the ground-truth is larger than a threshold \AF{varied from 0 to 1} (Figure~\ref{fig:results}~(a)). 
We also use the normalized precision plot \cite{TrackingNet}, that \AF{reports}, \AF{for a variety of distance thresholds},
the percentage of bounding boxes whose center points are within a given normalized distance (in pixels) from the ground-truth (Figure \ref{fig:results} (b)). 
\AF{As summary measures, we report the success score (SS) \cite{OTB} and normalized precision scores (NPS) \cite{TrackingNet}, which are computed as the Area Under the Curve (AUC) of the success plot and normalized precision plot respectively.}

Along with these standard metrics, we employ a novel plot which we refer to as generalized success robustness plot (Figure \ref{fig:results} (c)). We take inspiration from the robustness metric proposed in \cite{VOT2020} which measures the normalized extent of a tracking sequence before a failure. But differently from \cite{VOT2020}, which uses a fixed overlap threshold to detect a collapse, we propose to use different thresholds ranging in [0, 0.5]. This allows to assess the length of tracking sequences for different application scenarios.
We consider 0.5 as the maximum threshold as higher overlaps are usually associated to positive predictions in many computer vision tasks. 
\AF{Similarly as \cite{OTB,TrackingNet}, we use the AUC of the generalized robustness plot to obtain an aggregate score which we refer to as generalized success robustness (GSR).}
This new measure evaluates trackers' capability of maintaining long temporal reference to targets. We think this aspect is especially important in FPV as longer references to the target can lead to a better modeling of the camera viewer's actions and interactions with objects.

Finally, we evaluate the trackers' processing speed in frames per second (FPS) to quantify their efficiency.

\begin{figure*}[t]%
\centering
\includegraphics[width=\linewidth]{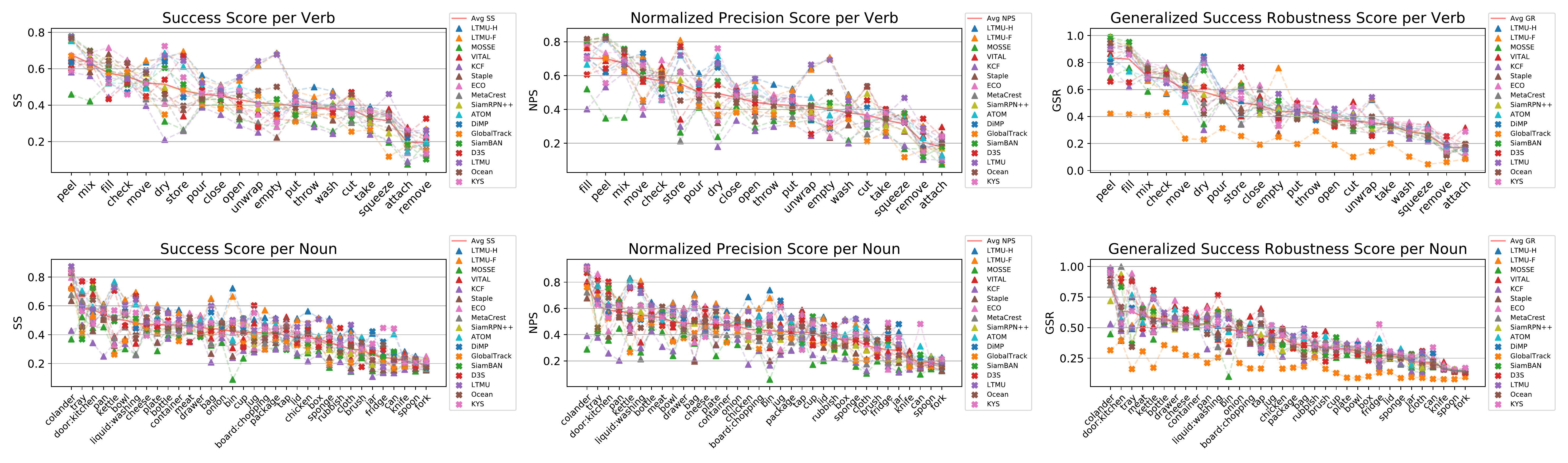}
\caption{SS, NPS, and GSR performance of 17 among the 33 selected trackers with respect to the action verbs (first row of plots) and target nouns (second row of plots) in \datasetname. The red plain line highlights the average performance.}
\label{fig:resverbsnouns}
\end{figure*}

\section{Results}
\label{sec:resuluts}

\begin{table*}[t]
\fontsize{6}{7}\selectfont
	\centering
	\caption{Performance achieved by 17 of the benchmarked trackers on \datasetname\ using the RTE protocol.}
	\label{tab:realtime}
	\setlength\tabcolsep{.15cm}
	\begin{tabular}{l | c c c c c c c c c c c c c c c c c }
		\toprule
		
		Metric & Ocean & SiamBAN & SiamRPN++ & DiMP & KYS & ATOM & LTMU & D3S & ECO & GlobalTrack & Staple & MOSSE & LTMU-H & MetaCrest & LTMU-F & VITAL & KCF \\

		\midrule
		
		FPS & 21 & 24 & 23 & 16 & 12 & 15 & 8 & 16 & 15 & 8 & 13 & 26 & 4 & 8 & 4 & 4 & 6 \\
		SS & 0.365 & 0.360 & 0.362 & 0.336 & 0.327 & 0.319 & 0.284 & 0.276 & 0.252 & 0.253 & 0.249 &  0.227 & 0.213 & 0.207 & 0.205 & 0.204 & 0.186 \\
		NPS & 0.358 & 0.366 & 0.356 & 0.331 & 0.317 & 0.312 & 0.257 & 0.263 & 0.231 & 0.227 & 0.236 & 0.190 & 0.174 & 0.175 & 0.161 & 0.165 & 0.157 \\
		GSR & 0.294 & 0.313 & 0.293 & 0.224 & 0.237 & 0.179 & 0.169 & 0.182 & 0.173 & 0.139 & 0.169 &  0.141 & 0.161 & 0.165 & 0.162 & 0.158 & 0.177 \\

		\bottomrule		
\end{tabular}
\end{table*}

\begin{table*}[t]
\fontsize{6}{7}\selectfont
	\centering
	\caption{Accuracy results on \datasetname\ of a video-based hand-object detection solution which considers each of the considered trackers as localization method for the object involved in the interaction. As a baseline, we employ the object detection capabilities of the hand-object interaction solution Hands-in-contact \cite{Shan2020}. }
	\label{tab:handsobj}
	\setlength\tabcolsep{.11cm}
	\begin{tabular}{c | c c c c c c c c c c c c c c c c c }
		\toprule
		Hands-in-contact \cite{Shan2020} & LTMU-H & LTMU-F & ATOM & LTMU & Ocean & SiamBAN & SiamRPN++ & MetaCrest & D3S & DiMP & KYS & VITAL & GlobalTrack & MOSSE & ECO & Staple & KCF \\
		\midrule
		0.354 & 0.368 & 0.367 & 0.361 & 0.354 & 0.340 & 0.340 & 0.311 & 0.293 & 0.292 & 0.292 & 0.279 & 0.253 & 0.251 & 0.231 & 0.230 & 0.197 & 0.177 \\

		\bottomrule		
\end{tabular}
\vspace{-0.5em}
\end{table*}

\pgraph{How Do the Trackers Perform in the FPV Scenario?}
Figure \ref{fig:results} \AF{reports} the performance of the selected trackers on \datasetname\ using the OPE protocol. \AF{For reference, we also report the performance of the trackers on the popular OTB-100 \cite{OTB} benchmark (semi-transparent curves - gray numbers in brackets).}
\AF{It can be clearly noted that} the overall performance \AF{of the trackers} is decreased across all measures \AF{when considering the challenging FPV scenario of \datasetname}. 
For example, the \AF{SS, NPS, and GSR scores of} LTMU 
on \datasetname\ \AF{are} 
 43.7\% , 44.8\%, and 43.1\%,
 which \AF{are} much lower than the respective 69.6\%, 76\%, and 78\%, achieved on OTB-100. 
 With the MSE protocol, LTMU achieves the respective scores of 46.9\%, 48.3\%, 38.6\%.\footnote{See Appendix \ref{sec:addres} for the overall MSE results of all trackers.}
These results show that the particular characteristics of FPV present in \datasetname\ introduce challenging scenarios for visual trackers.
Some qualitative examples of the trackers' performance are shown in Figure \ref{fig:qualitative} of the Appendix.

\AF{Generally speaking,}
trackers \AF{based on deep learning} (e.g. LTMU, TRASFUST, ATOM, KYS, Ocean) %
perform better in SS and NPS than \AF{those based on}
hand-crafted features (e.g. BACF, MCCTH, DSST, KCF).
Among the first \AF{class of} trackers, the ones leveraging online adaptation mechanisms (e.g. LTMU, ATOM, VITAL, ECO, KYS, DiMP) are more accurate than \AF{the ones based on} single-shot instances (e.g. Ocean, D3S, SiamRPN++).
The generalized success robustness plot in Figure \ref{fig:results}(c) and the GSR results of Figure \ref{fig:resultsmse} of the supplementary report a different rankings of the trackers, showing that more spatially accurate trackers are not always able to maintain longer reference to targets.

Under both the OPE and MSE protocols, the proposed FPV trackers LTMU-H and LTMU-F are largely better in SS and NPS, while they lose some performance in GSR.
Such outcome shows that adapting a state-of-the-art method to FPV allows to marginally improve results, while we expect significant performance improvements to be achievable by a tracker accurately designed to tackle the FPV challenges introduced by this benchmark.

\pgraph{In Which Conditions Do the Trackers Work Better?}
Figure \ref{fig:resultsattributes} \AF{reports} the SS, NPS, and GSR scores, computed with the MSE protocol, of 17 trackers with respect to the 
attributes introduced in Table \ref{tab:attrdesc}.\footnote{\label{note1}The analysis was restricted to 17 trackers for a better visualization of the plots/tables. The 17 trackers were selected to represent various methodologies. The results for all trackers are available in Appendix \ref{sec:addres}.} 
We do not report results for the POC \AF{attribute} as it is present in every sequence, as shown in Figure \ref{fig:distributions}~(a).
It stands out clearly that full occlusion~(FOC), out of view~(OUT) and the small size of targets~(LR) \AF{are} the most difficult situations for trackers. The fast motion of targets~(FM) and the presence of similar objects~(SOB) are also critical factors that cause drops in performance. 
Trackers show to be less vulnerable to rotations~(ROT) and to the illumination variation~(IV). 
Generally, tracking rigid objects~(RIG) results easier than tracking deformable ones (DEF).
With respect to the new \AF{4} sequence attributes related to FPV, it results that tracking objects held with two hands~(2H) is more difficult than tracking objects held with a single hand~(1H). This is probably due to the \AF{additional occlusions generated in the 2H scenario.}
Trackers are instead quite robust to \AF{head motion}~(HM)
\AF{and seem to cope better with objects appearing in larger size~(HR).}

\pgraph{How Do the Trackers Perform With Respect to the Actions?}
The first row of plots in Figure \ref{fig:resverbsnouns} 
\AF{reports}
the MSE protocol results of \AF{SS, NPS, and GSR with respect to the associated verb action labels.}$^{\ref{note1}}$
Actions that \AF{mainly} cause a spatial \AF{displacement} 
of the target (e.g. ``move'', ``store'', ``check'') generally have less impact on the
performance.
Actions that change the state, shape, or aspect ratio of an object (e.g. ``remove'', ``squeeze'', ``cut'', ``attach'') generate harder tracking scenarios.
\AF{Also the sequences characterized by the ``wash'' verb lead trackers to poor performance.}
Indeed, 
\AF{the wash action}
can cause many occlusions
\AF{and}
make the object harder to track.

The second row of the same figure presents the performance scores of the trackers with respect to the associated noun labels.
\AF{Rigid, regular-sized objects such as ``pan'', ``kettle'', ``bowl'', ``plate'', and ``bottle'' are among the ones associated with high average scores.}
On the other hand, \AF{some rigid objects such as ``knife'', ``spoon'', ``fork'' and ``can'' are harder}
to track, probably due to their particularly thin shape and the light reflectance they are \AF{easily} subject to.  
\AF{Deformable objects such as ``sponge'', ``onion'',  ``cloth'' and ``rubbish'' are in general also difficult to track.}

\pgraph{How Fast Are the Trackers?}
Table \ref{tab:realtime} \AF{reports} the FPS performance of the trackers and the SS, NPS, and GSR scores achieved under the RTE protocol.$^{\ref{note1}}$
None of the trackers achieve the frame rate speed of 60 FPS. We argue \AF{that} this is due the full HD \AF{resolution of} frames \AF{which requires demanding image crop and resize operations with targets of considerable size.}
Thanks to their \AF{non-reliance of}
online adaptation mechanisms, \AF{trackers based on siamese networks} (e.g. Ocean, SiamBAN, SiamRPN++)
emerge as the fastest trackers and exhibit a less significant performance drop of the proposed scores. %
Trackers using online learning approaches (e.g. ATOM, DiMP, ECO, KYS) 
generally \AF{achieve a below real-time} %
speed, consequently causing a major accuracy loss when deployed to real-time scenarios. %
In general, we observe that the GSR score is the measure on which all trackers present the major  drop in the real-time setting, suggesting that particular effort should be spent to better model actions and interactions in such scenarios.

\pgraph{Do Trackers Already Offer Any Advantage in FPV?}
Despite we are demonstrating that FPV is challenging for current trackers, \AF{we assess whether these already offer an advantage in the FPV domain to obtain information about the objects' locations and movements in the scene~\cite{Wang2020,Furnari2017,RULSTMpami,sener2020temporal,Shan2020}.
To this aim, we performed two experiments.\footnote{Details are given in the Appendix \ref{sec:expdetail} of the supplementary material.}
First, we evaluated the performance of a Faster R-CNN~\cite{FasterRCNN} instance trained on EK-55 \cite{EK55} when used as a naive tracking baseline.} Such a solution achieves an SS, NPS, and GSR of 0.323, 0.369, 0.044, by running at 1 FPS. Comparing these results with the ones presented in Figure \ref{fig:results}, we clearly notice that trackers, if properly initialized by a detection module, can deliver faster, more accurate and much more temporally long object localization than detectors.

As a second experiment, we evaluated the accuracy of a video-based hand-object interaction detection solution \cite{Shan2020} whose object localisation is given by a tracker rather than a detector. The tracker is initialized with the object detector's predicted bounding-box at the first detection of the hand-object interaction, and let run until its end. By this setting, we created a ranking of the trackers which is presented in Table \ref{tab:handsobj}. The results demonstrate that stronger trackers can improve the accuracy and efficiency of current detection-based methodologies \cite{Shan2020}. Interestingly, the trackers' ranking differs from what shown in Figure \ref{fig:results}, suggesting that trackers can manifest other capabilities when deployed into application scenarios.

\AF{Given these preliminary results, we hence expect that trackers will likely gain more importance in FPV as new methodologies explicitly considering the first person point of view are investigated.}

\section{Conclusions}
In this paper, we \AF{proposed} the first systematic evaluation of visual object tracking in FPV. The analysis \AF{has been} conducted with standard and novel measures on the newly introduced \datasetname\ benchmark, \AF{which} contains 150 video sequences extracted from the EK-55 \cite{EK55} FPV dataset. \datasetname\ has been densely annotated with 97K bounding-boxes, 17 sequence attributes, 
\AF{20 action verb attributes and 34 target object attributes.}
The performance of 31 state-of-the-art visual trackers and two baseline FPV trackers was analysed extensively on the proposed dataset. The results show a generalized drop in accuracy with respect to the \AF{performance} achieved on existing tracking benchmarks. \AF{Furthermore, our analysis provided} insights about which scenarios and actions cause the performance to change. Finally, we \AF{have shown} that object tracking gives an advantage in terms of \AF{object localization} accuracy and efficiency \AF{over} object detection.
\AF{These results suggest that FPV is a challenging scenario for current trackers and that tracking will likely get more importance in this domain as new FPV-specific solutions will be investigated.}
Annotations, results, and code, are available at \datasetlink.

{\footnotesize
\noindent\textbf{Acknowledgements.}
 Research at the University of Udine
has been supported by the ACHIEVE-ITN H2020 project.
Research at the University of Catania has been supported by
MIUR AIM - Attrazione e Mobilita Internazionale Linea 1
- AIM1893589 - CUP: E64118002540007.}

{\small
\bibliographystyle{ieee_fullname}
\bibliography{egbib}
}

\clearpage

\appendix

\begin{figure*}[t]%
\centering
\includegraphics[width=\linewidth]{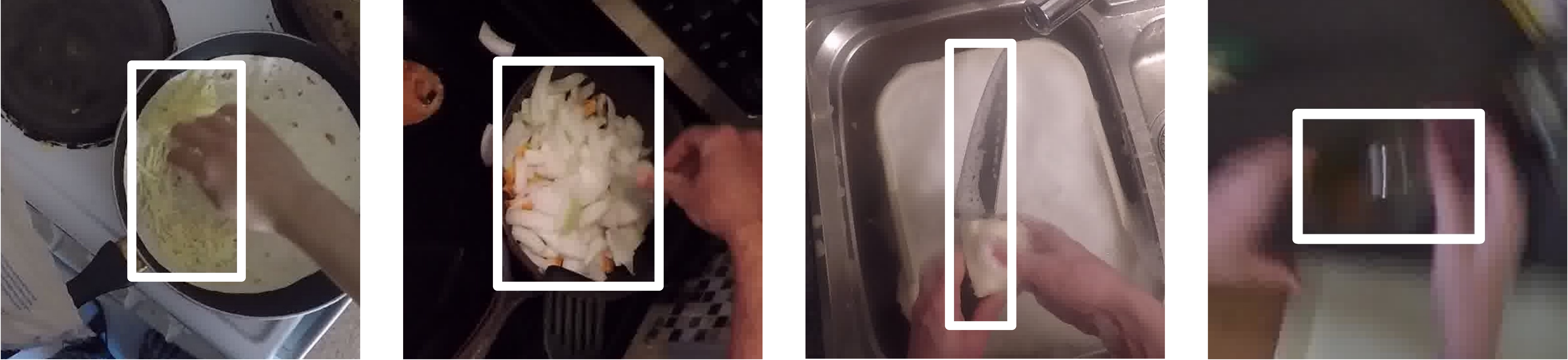}
\caption{Examples of target objects contained in \datasetname\ that are difficult to represent with more sophisticated representations (e.g. rotated bounding box or segmentation mask). The first two images from the left show objects such as ``cheese'' and ``onion'' which prevent the determination of the angle for an oriented bounding box, or an accurate segmentation mask. The last two images present objects which prevent a consistent definition of a segmentation.}
\label{fig:hardex}
\end{figure*}

\section{Motivations And Details Behind \datasetname}
\label{sec:datasetdiff}
In this section, we provide more motivations and details behind the construction of the \datasetname\ benchmark dataset. 

First of all, we remark that \datasetname\ has been designed for the \emph{evaluation} of visual tracking algorithms in FPV regardless of their methodology. Indeed, this paper does not aim to provide a large-scale dataset to improve the performance of deep learning based trackers. Instead, its goal is to assess the impact of the first-person viewpoint on current trackers and, to the best of our knowledge, this analysis was never done before. Hence, as a \emph{first step} towards providing an answer to such a point (which is also highlighted in the title of the paper), we focused on benchmarking the tracking progress made by the computer vision community in the last years.

\pgraph{Video Collection.}
The video sequences contained in \datasetname\ have been sampled from the EPIC-Kitchens-55 (EK-55) dataset \cite{EK55}. This has been done because EK-55 is currently the largest dataset for understanding human-object interactions in FPV (it provides up to 55 hours of human-object interaction examples). Thanks to its dimension, it is the only database that provides a significant amount of diverse interaction situations between various people and several different types of objects. Hence, it allowed us to select suitable diverse tracking sequences that reflect the common scenarios tackled in FPV tasks.

\pgraph{Bounding Box Annotations.} 
To represent the spatial localization of objects, we employed axis-aligned bounding-boxes. This design choice for the \datasetname\ benchmark is supported by the fact that this representation is largely used in many FPV pipelines \cite{Furnari2017,RULSTMiccv,RULSTMpami,EK55,Kapidis2019,Visee2020,Shan2020}. Therefore, computing performance results based on such allows us to correlate them to the results of other FPV tasks that employ the same object representation. Hence, we can better highlight the impact that trackers would have in such contexts.
Moreover, we would like to highlight the difficulty that the FPV setting poses on the development of more sophisticated annotations for object categories that appear commonly in FPV scenarios. Figure \ref{fig:hardex} shows some examples of these. The first two images from the left show the objects ``cheese'' and ``onion'' (these are considered as single objects according to the EK-55 annotations \cite{EK55}) which prevent the determination of the angle for an oriented bounding-box, or an even accurate segmentation mask due to their spatial sparsity. The two images on the right present objects for which providing a segmentation is very ambiguous. Indeed, most of the pixels in the image area of the knife (third image) belong actually to foam, while the heavy motion blur happening on the object of the fourth image (where the target is a bottle) prevents the definition of the actual pixels belonging to the object. In all these scenarios, axis-aligned bounding-boxes result in robust target representations that provide a consistent delineation of the object. For these motivations, and to make representations and annotations consistent across the whole dataset, we employed such annotation representations.

Moreover, the latest progress of visual tracking algorithms on various benchmarks that use this state representation \cite{OTB,UAV123,NfS,CDTB,TrackingNet,LaSOT,GOT10k} demonstrates that it provides sufficient information about the target for consistent and reliable performance evaluation. Furthermore, using more sophisticated target representation would have restricted our analysis \cite{SiamMask,D3S,Dunnhofer2020eccvw,LWL} since the majority of state-of-the-art trackers output just axis-aligned bounding boxes \cite{MOSSE,DSST,KCF,MDNet,Staple,SiamFC,GOTURN,ECO,BACF,VITAL,STRCF,MCCTH,MetaTrackers,SiamRPNpp,SiamDW,ATOM,DiMP,SPLT,PrDiMP,GlobalTrack,SiamFCpp,LTMU,SiamBAN,Ocean,KYS}.

\begin{figure}[t]%
\centering
\includegraphics[width=\columnwidth]{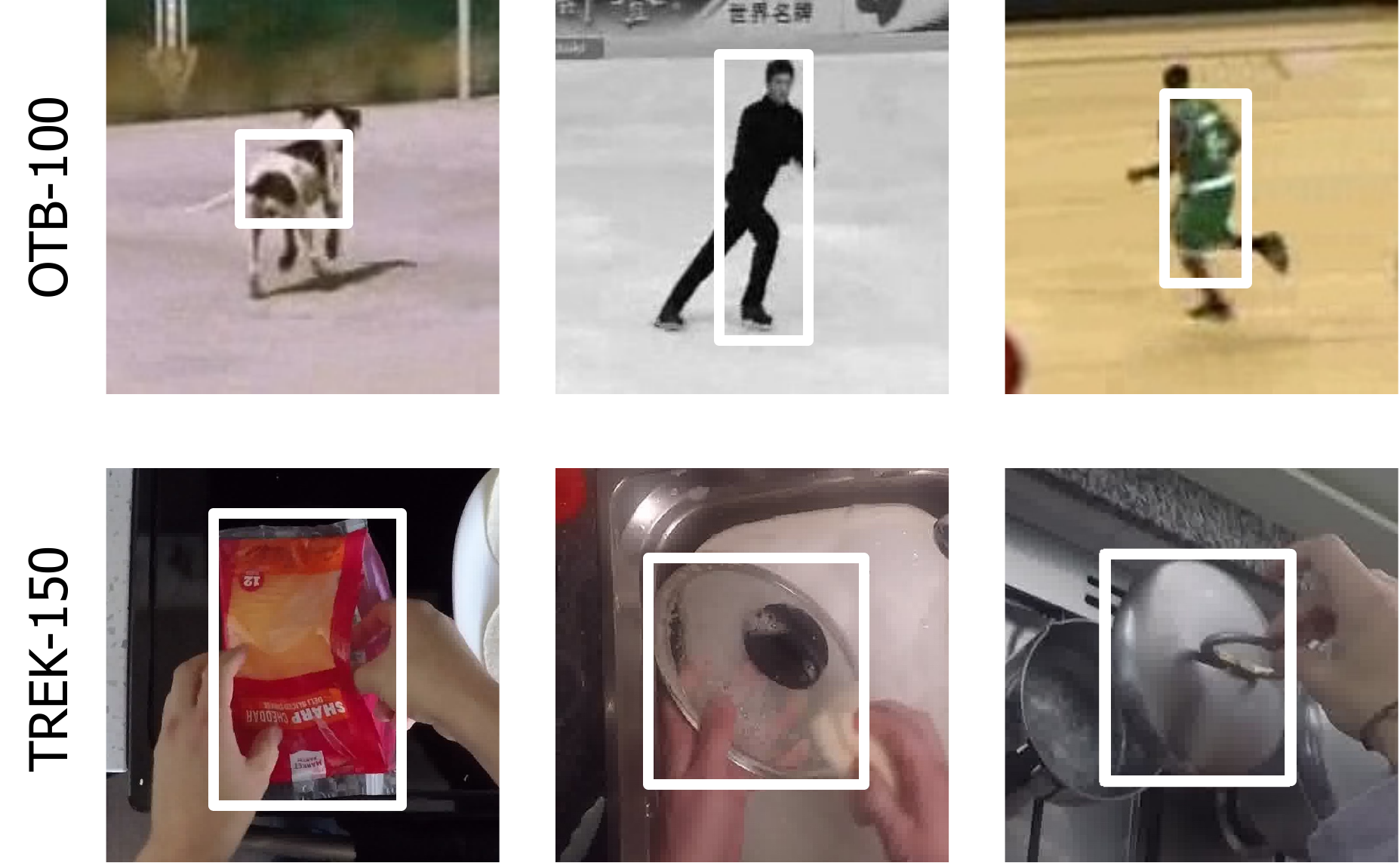}
\caption{Examples of the quality of the bounding box annotations contained in \datasetname\ in comparison with the ones available in the popular OTB-100 benchmark. \datasetname\ provides careful and high-quality annotations that tightly enclose all the target objects.}
\label{fig:annoquality}
\end{figure}

Finally, we point out that the proposed axis-aligned bounding-boxes have been carefully and tightly drawn around the visible parts of the objects. Figure \ref{fig:annoquality} shows some examples of the quality of the bounding box annotations of \datasetname\ in contrast to the ones available in the popular OTB-100 tracking benchmark.

\pgraph{Frame Rate.}
The videos contained in \datasetname\ have a frame rate of 60 FPS. This is inherited from the EK-55 dataset \cite{EK55},  from which videos are sampled. According to the authors \cite{EK55}, EK-55 has been acquired with such a setting because of the proximity of the camera point of view and the main scene (i.e. manipulated objects), which causes very fast motion and heavy motion blur when the camera wearer moves (especially when he/she moves the head). 

We empirically evaluated the fast motion issue by assessing the average normalized motion happening on the frames that include fast motion (FM) (we computed them by considering the automatic procedure defined in \cite{OTB,TrackingNet} to assign the FM attribute). Such a motion quantity has been computed as the distance between the center of two consecutive ground-truth bounding boxes normalized by the frame size.
Considering \datasetname\ with the videos at 30 FPS, such a value achieves 0.075. This is higher than the 0.068 obtained for OTB-100, the 0.033 of UAV123, or the 0.049 of NfS considered at 30 FPS. These comparisons demonstrate that the FPV scenario effectively includes challenging scenarios due to the faster motion of targets/scene. Considering the 60 FPS frame rate, the fast motion quantity of \datasetname\ is reduced to 0.062, which is comparable to the values obtained in other third-person tracking benchmarks.

\pgraph{Sequence Labels.}
To study the performance of trackers under different aspects, the sequences of \datasetname\ have been associated with one or more of 17 attributes that indicate the visual variability of the target in the sequence (see Table \ref{tab:attrdesc} of the main paper for the details). The extended usage of this practice \cite{OTB,UAV123,NfS,NUSPRO,TrackingNet,LaSOT,GOT10k} showed how this kind of labeling is sufficient to estimate the trackers' performance on particular scenarios. 
We therefore follow such an approach to associate labels on \datasetname's videos. 
However, we argue that, by using this labeling setting, attention must be paid to how trackers are evaluated. The standard OPE protocol, which has been generally used to perform such evaluations, could lead to less accurate estimates. For example, it could happen that a tracker would fail for some event described by an attribute (e.g. FOC) in the first frames of a video, but that the sequence also contains some other event (e.g. MB) in the end. With the score averaging procedure defined by the OPE protocol, the low results achieved due to the first event would set low scores also for the second event, while the tracker failed just for the first one. Therefore, the performance estimate for the second attribute would not be realistic. 
We believe a reasonable option is to use a more robust evaluation protocol such as the multi-start evaluation (MSE).
Thanks to its points of initialization which generate multiple diverse sub-sequences, this protocol allows a tracker to better cover all the possible situations happening along the videos, both forward and backward in time. All the results achieved on the sub-sequences are then averaged to obtain the overall scores on a sequence. We think the scores computed in this way to be more robust and accurate estimates of the real performance of the trackers. Hence, in this work, we follow such an approach to evaluate trackers over sequence attributes.

\begin{figure*}[t]%
\centering
\includegraphics[width=\linewidth]{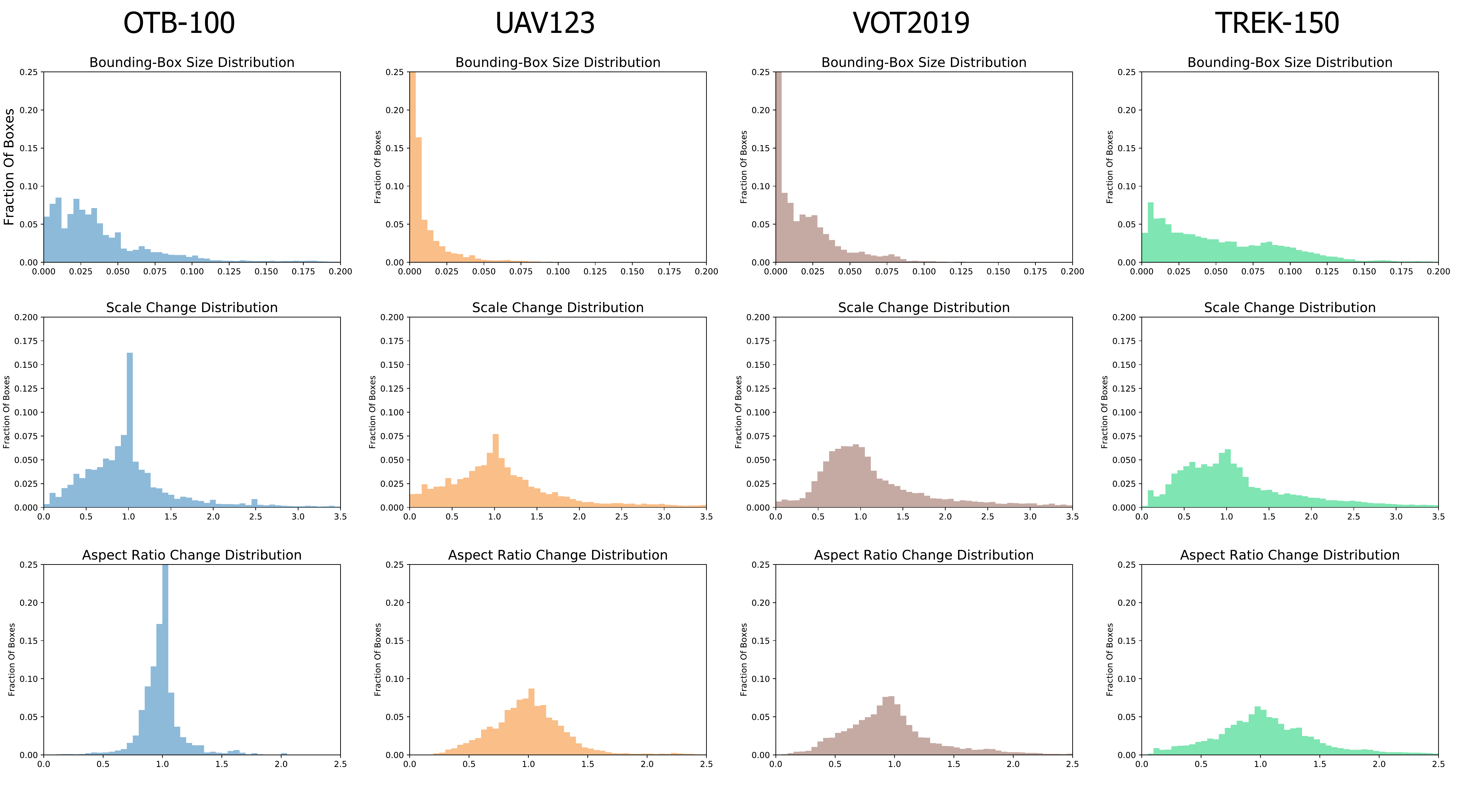}
\caption{Comparison between \datasetname\ (last column of plots) and other popular visual tracking benchmarks on the distributions computed for different bounding box characteristics. Each column of plots reports the distribution of bounding box sizes, scale changes, and aspect ratio change (the x-axis of each plot reports the range of the bounding box statistic).}
\label{fig:boxdist}
\end{figure*}

\pgraph{Single Object Tracking.}
In this paper, we restricted our analysis to the tracking of a single object per video. This has been done because in the FPV scenario a person interacts through hands with one or two objects at a time in general \cite{EK55} (if a person interacts with two objects they can be still tracked by two single-object trackers). Moreover, focusing on a single object allows us to analyze better all the challenging and relevant factors that characterize the tracking problem in FPV.
We believe that future work could investigate the employment of multiple object tracking (MOT) solutions \cite{MOT} for a general understanding of the position and movement of all objects visible in the scene. We think that the study presented in this paper will give useful insights even for the development of such methods.

\pgraph{Differences With Other Tracking Benchmarks.}
We believe that the proposed \datasetname\ benchmark dataset offers \emph{complementary} features with respect to existing visual tracking benchmarks.

Table \ref{tab:datasets} and Figure \ref{fig:distributions}(a) and (b) of the main paper show that \datasetname\ provides complementary characteristics to what is available today to study the performance of visual trackers. Particularly, our proposed dataset offers different distributions of the common challenging factors encountered in other datasets. For example, \datasetname\ includes a larger number of examples with occlusions (POC), fast motion (FM), scale change (SC), aspect ratio change (ARC), illumination variation (IV), and motion blur (MB), while it provides a competitive number of scenarios for low resolution (LR), full occlusion (FOC), deformable objects (DEF), and presence of similar objects (SOB). Additionally, even though the 4 new attributes high resolution (HR), head motion (HM), one-hand interaction (1H),  two-hands interaction (2H), define particular FPV scenarios, we think that they can be of interest even for the visual tracking community. For example, as shown by the second row of images of Figure \ref{fig:annoquality}, 1H and 2H can be considered as attributes that define different levels of occlusion, as objects manipulated with two hands generally cause more extended hiding of the targets. 
Besides these sequence-level features, \datasetname\ offers up to 34 target categories which, to the best of our knowledge, have never been studied. As shown by the Figures \ref{fig:hardex} and \ref{fig:annoquality}, these objects have challenging appearances (e.g. transparent or reflective objects like lids, bottles, or food boxes) and shapes (e.g. knives, spoons, cut food) that change dramatically due to the interaction or motion induced by the camera viewer.

We additionally computed some statistics on the bounding box ground-truth trajectories contained in the proposed dataset. Figure \ref{fig:boxdist} reports these distributions. For comparison, we report the distributions computed on the popular tracking benchmarks VOT2019, UAV123, OTB-100. As can be noted, our dataset exhibits different distributions, and thus offers different behaviors of the target appearances and motions. Particularly, observing the first plot of the last column, it can be noted that \datasetname\ has a wider distribution of bounding box sizes, hence making it suitable for the evaluation of trackers with targets of many different sizes. Particularly, \datasetname\ has a larger number of bounding boxes with greater dimension. The plot just below the first shows that \datasetname\ provides more references to assess the trackers' capabilities in tracking objects that become smaller. Finally, the last plot shows a wider distribution for the aspect ratio change, showing that \datasetname\ offers a large variety of examples to evaluate the capabilities of trackers in predicting the shape change of targets.

Additionally to these characteristics, we think \datasetname\ is interesting because it allows the study of visual object tracking in unconstrained scenarios of \emph{every-day} situations.

\begin{table*}[t]
\fontsize{9}{10}\selectfont
	\centering
	\caption{Details of the trackers involved in our evaluation. In the Image Representation column the acronyms stand for: CNN - Convolutional Neural Network; HOG - Histogram of Oriented Gradients; Pixel - Pixel Intensity; Color - Color Names or Intensity.  Regarding the Matching strategy column the acronyms stand for: CF - Correlation Filter; CC - Cross Correlation; T-by-D - Tracking by Detection; Reg - Regression; Had - Hadamard Correlation. The \cmark\ symbol in the Model Update column expresses the target model update during the tracking procedure. The last column reports the tracking method class according to \cite{Lukezic2018me} ($\text{ST}$ - Short-Term trackers, $\text{LT}$ - Long-Term trackers). }
	\label{tab:trackers}
	\begin{tabular}{l | c  | c  c  c  c }
		\toprule
		Tracker & Venue & Image Representation & Matching & Model Update &  \cite{Lukezic2018me} Class \\
		\midrule
		MOSSE \cite{MOSSE} & CVPR 2010 & Pixel & CF & \cmark & $\text{ST}_0$ \\
		DSST \cite{DSST} & BMVC 2014 & HOG+Pixel & CF & \cmark & $\text{ST}_0$ \\
		KCF \cite{KCF} & TPAMI 2015 & HOG & CF & \cmark & $\text{ST}_0$ \\
		MDNet \cite{MDNet} & CVPR 2016 & CNN & T-by-D & \cmark & $\text{ST}_1$ \\
		Staple \cite{Staple} & CVPR 2016 & HOG+Color & CF & \cmark & $\text{ST}_0$ \\
		SiamFC \cite{SiamFC} & ECCVW 2016 & CNN & CC & \xmark & $\text{ST}_0$ \\
		GOTURN \cite{GOTURN} & ECCV 2016 & CNN & Reg & \xmark & $\text{ST}_0$ \\
		ECO \cite{ECO} & CVPR 2017 & CNN & CF & \cmark & $\text{ST}_0$ \\
		BACF \cite{BACF} & ICCV 2017 & HOG & CF & \cmark & $\text{ST}_0$ \\
		DCFNet \cite{DCFNet} & ArXiv 2017 & CNN & CF & \cmark & $\text{ST}_0$ \\
		VITAL \cite{VITAL} & CVPR 2018 & CNN & T-by-D & \cmark & $\text{ST}_1$ \\
		STRCF \cite{STRCF} & CVPR 2018 & HOG & CF & \cmark & $\text{ST}_0$ \\
		MCCTH \cite{MCCTH} & CVPR 2018 & HOG +Color & CF & \cmark & $\text{ST}_0$ \\
		DSLT \cite{DSLT} & ECCV 2018 & CNN & CC & \cmark & $\text{ST}_0$ \\
		MetaCrest \cite{MetaTrackers} & ECCV 2018 & CNN & CF & \cmark  & $\text{ST}_1$ \\
		SiamRPN++ \cite{SiamRPNpp} & CVPR 2019 & CNN & CC & \xmark & $\text{ST}_0$ \\
		SiamMask \cite{SiamMask} & CVPR 2019 & CNN & CC & \xmark & $\text{ST}_0$  \\
		SiamDW \cite{SiamDW} & CVPR 2019 & CNN & CC & \xmark & $\text{ST}_0$  \\
		ATOM \cite{ATOM} & CVPR 2019 & CNN & CF & \cmark & $\text{ST}_1$ \\
		DiMP \cite{DiMP} & ICCV 2019 & CNN & CF & \cmark & $\text{ST}_1$ \\
		SPLT \cite{SPLT} & ICCV 2019 & CNN & CF & \cmark & $\text{LT}_1$ \\
		UpdateNet \cite{UpdateNet} & ICCV 2019 & CNN & CC & \cmark & $\text{ST}_0$ \\
		SiamFC++ \cite{SiamFCpp} & AAAI 2020 & CNN & CC & \xmark & $\text{ST}_0$ \\
		GlobalTrack \cite{GlobalTrack} & AAAI 2020 & CNN & Had & \xmark & $\text{LT}_0$\\
		PrDiMP \cite{PrDiMP} & CVPR 2020 & CNN & CF & \cmark & $\text{ST}_1$ \\
		SiamBAN \cite{SiamBAN} & CVPR 2020 & CNN & CC & \xmark & $\text{ST}_0$ \\
		D3S \cite{D3S} & CVPR 2020 & CNN & CF & \xmark & $\text{ST}_0$ \\
		LTMU \cite{LTMU} & CVPR 2020 & CNN & CF/CC & \cmark & $\text{LT}_1$ \\
		Ocean \cite{Ocean} & ECCV 2020 & CNN & CC & \xmark & $\text{ST}_0$ \\
		KYS \cite{KYS} & ECCV 2020 & CNN & CF & \cmark & $\text{ST}_1$ \\
		TRASFUST \cite{Dunnhofer2020accv} & ACCV 2020 & CNN & Reg & \xmark & $\text{ST}_1$ \\
		\bottomrule		
\end{tabular}
\end{table*}

\section{Tracker Details}
\label{sec:trackerdetails}

\pgraph{Generic Object Trackers Details.}
Table \ref{tab:trackers} reports some additional information about the 31 considered generic-object trackers such as: venue and year of publication; type of image representation used; type of matching strategy; employment of target model updates; and category of tracker according to the classification of~\cite{Lukezic2018me}. For each tracker, we used the code publicly available and adopted default parameters for evaluation purposes.

\pgraph{FPV Trackers Details.}
In this section, we provide details on the LTMU-F and LTMU-H FPV trackers considered as baselines in our study. For a better understanding, we briefly recap the processing procedure of the LTMU tracker~\cite{LTMU}. After being initialized with the target in the first frame of a sequence, at every other frame LTMU first executes the short-term tracker DiMP~\cite{DiMP} that tracks the target in a local area (based on the target's last known position) of the frame. The image patch extracted from the bounding box prediction of DiMP is evaluated by an online-learned verifying module which outputs a probability estimate  for the target being contained in the patch. Such an estimate is employed to decide if the short-term tracker is tracking the target or not. If it is, the box predicted by the short-term tracker is given as output for the current frame. In the other case, a re-detection module is executed to search for the target in the global frame. The detector returns some candidate locations to contain the target and each of these is checked by the verification module. The candidate patch with the highest confidence is given as output and used as a new target location to reset the short-term tracker. 

\begin{figure*}[t]%
\centering
\includegraphics[width=\linewidth]{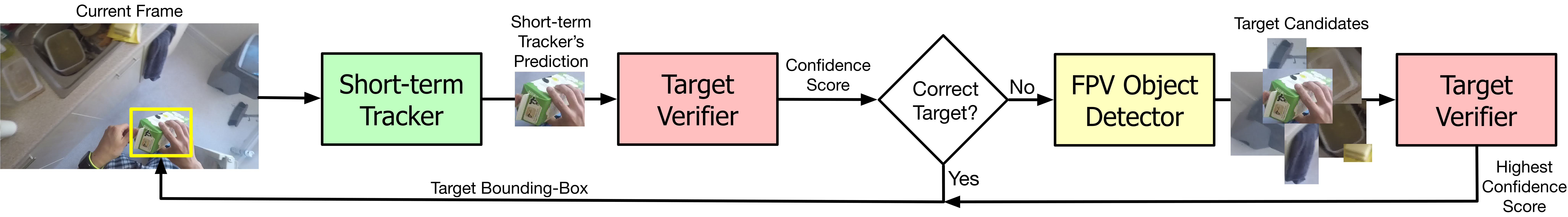}
\caption{Visual representation of the LTMU \cite{LTMU} scheme performed at every frame that has been adapted for the development of the baseline FPV trackers LTMU-F and LTMU-H.}
\label{fig:fpvtracker}
\end{figure*}

In our setting, we employ FPV-based detectors to implement such a re-detection module. For LTMU-F, we employed the EK-55 trained Faster R-CNN \cite{EK55}. Among the many detections given as output, this module has been set to retain the first 10, considering a ranking based on the scores attributed by the detector to each detection. If no detection is given for a frame, the last available position of the target is considered as candidate location. For LTMU-H, we employ the object localization contained in the hand-object interaction detections given by the FPV version of Hands-in-contact \cite{Shan2020} to obtain the target candidate locations. Such a solution \cite{Shan2020} is implemented as an improved Faster R-CNN which is set to learn to provide, at the same time, the localization of hands and objects, and their state of interaction. As before, if no detection is given for a frame, the last available position of the target is considered as candidate location.
For both methods, the original pre-trained models (made available by the authors) which consider FPV data have been used.
The described setups, which the common scheme is presented in Figure \ref{fig:fpvtracker}, give birth to two trackers that implement conceptually different strategies for FPV-based object localization. Indeed, the first solution reasons just to find objects in the scene, while the second reasons in terms of the interaction happening between the camera wearer (i.e. hands) and the objects. We would like to remark that other FPV trackers (such as the ones described in Section \ref{sec:fpvtrackers} of the main paper) have not been tested on \datasetname\ because their implementations are not available.

\pgraph{Implementation Details.}
The evaluations were performed on a machine with an Intel Xeon E5-2690 v4 @ 2.60GHz CPU, 320 GB of RAM, and an NVIDIA TITAN V GPU.
\AF{We considered the Python publicly available implementations of each tracker and adopted default parameters.}
Annotations, results of the trackers, and code are available at \datasetlink.

\section{Experimental Details}
\label{sec:expdetail}

\pgraph{Details On The Generalized Robustness.}

The robustness measure has been first introduced in \cite{Kristan2016}. This metric was defined as the number of drifts (i.e. the complete non-overlap between predictions and ground-truths) performed by a visual tracking algorithm.
In the last iteration of the VOT challenge \cite{VOT2020}, such a robustness measure has been revised and defined as the extent of a tracking sequence before the tracker's failure. Such an extent is determined as the number of frames positively tracked normalized by the total number of frames in the sequence. The failure event is triggered when the overlap between the predicted and ground-truth bounding-boxes becomes lower than a fixed threshold (the value 0.1 is used in \cite{VOT2020}). In simpler words, this measure expresses the fraction of a tracking sequence that is correctly tracked from its beginning.
We think this measure is of special interest to the FPV community. Indeed, it can assess the ability of a tracker to maintain reference in time to the target objects. 
Since many FPV tasks are devoted to understand the action performed by the camera viewer or its interaction with objects \cite{Furnari2017,RULSTMiccv,RULSTMpami,EK55,ragusa2020meccano}, having solutions capable of maintaining temporally longer references to target nouns can be advantageous to model such events.
However, we believe that having a single fixed threshold is restrictive, as different applications can make different assumptions on the concept of tracking failure. Therefore, following \cite{OTB} which proposed to evaluate trackers with plots computed after thresholding bounding-box overlaps with different values, we propose to build a plot considering different overlap thresholds for the determination of failure in the robustness measure \cite{VOT2020}. This leads to the creation of the Generalized Success Robustness Plot (Figure \ref{fig:results}(c) of the main paper) which reports the different robustness scores for the different thresholds. The latters have been studied just in the range [0, 0.5] because it is common practice, in the computer vision literature, to consider overlaps greater than 50\% as positive predictions. Notice that failures could be also defined in terms of center error. In this paper, we focused on overlap-based failures since bounding-box overlap has been shown to be superior for target localization accuracy \cite{Cehovin2016}, but future work will investigate the employment of the center error as this kind of bounding box distance is used in FPV tasks \cite{Shan2020}.
Moreover, to compare trackers with a single score, following \cite{OTB} and \cite{TrackingNet}, we compute the AUC of the Generalized Success Robustness Plot which we refer as to generalized success robustness (GSR). This value expresses the average of all the scores obtained with the different thresholds. In other words, the GSR score expresses the average successful extent of the predictions of a tracker.

\begin{figure*}[t]%
\centering
\includegraphics[width=\linewidth]{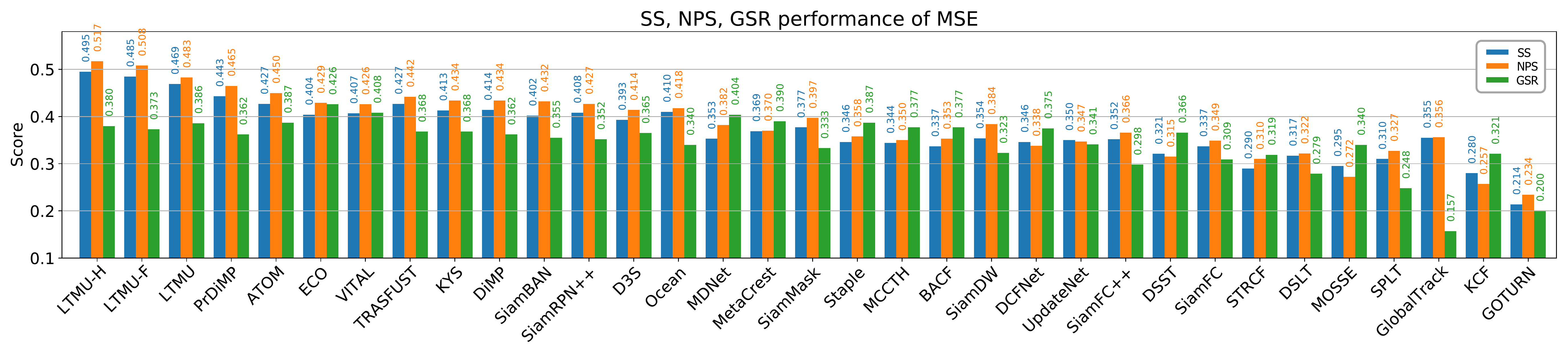}
\caption{SS, NPS, and GSR performance of the 33 benchmarked trackers on the proposed \datasetname\ benchmark under the MSE protocol. The general low performances confirms the conclusions achieved with the OPE protocol.}
\label{fig:resultsmse}
\end{figure*}

\pgraph{Details On The Evaluation Protocols.}
In this section, we give further details on the experimental protocols used for the execution of the trackers.

The one-pass evaluation (OPE) protocol, which is detailed in \cite{OTB}, consists of two main stages: (i) initializing a tracker with the ground-truth bounding box \AF{of} the target in the first frame; (ii) let the tracker run \AF{on}
 every \AF{subsequent} frame until the end of the sequence \AF{and record} predictions to be considered for the evaluation. For each sequence, predictions and ground-truth bounding boxes are compared according to the employed measures (only for frames where ground-truths are present) to obtain the performance scores. The overall scores, presented in brackets in Figure \ref{fig:results} of the main paper, are obtained by averaging the scores achieved for every sequence.

 For the implementation of the multi-start evaluation (MSE) protocol, we followed the details given in \cite{VOT2020}. For each sequence, different points of initialization (called anchors) separated by 2 seconds (in our setting every 120 frames) are defined. Anchors are always set in the first and last frame of a sequence. Some anchors are shifted forward for a few frames to obtain a more consistent bounding-box for tracker initialization. A tracker is run on each of the sub-sequences yielded by the anchor (in total 1032 sub-sequences are generated), either forward or backward in time depending on the longest sub-sequence the anchor generates. The tracker is initialized with the ground-truth in the first frame of the sub-sequence and let run until its end. Then, similarly as for the OPE, predicted and ground-truth bounding boxes are compared to obtain the performance scores for each sub-sequence. Scores for a single sequence are computed by a weighted average where the scores of each sub-sequence are weighted by its length (as number of frames). Similarly, the overall scores for the whole dataset (which are shown in Figure \ref{fig:resultsmse}) are obtained by a weighted average where each sequence's score is weighted by the number of frames in that sequence.

 The real-time evaluation (RTE) protocol has been implemented following the details in \cite{VOT2017,Li2020}. Similar to the OPE protocol, a tracker is initialized with the ground-truth in the first frame of a sequence. Then the tracker is presented with a new frame only after its execution over the previous frame has finished. The new presented frame is the last frame available for the time instant in which the tracker becomes ready to be executed, considering that frames occur regularly according to the frame rate of the video. In other words, all the frames occurring in the time interval between the start and end time instants of the tracker's execution are skipped. For all the frames skipped, the last bounding box given by the tracker is used as location for the target in such frames. The sequence and overall scores are ultimately obtained as for the OPE protocol.

\pgraph{Experiments On The Impact of Trackers In FPV.}
In this section, we report more details on the experiments performed to evaluate the impact of trackers in FPV tasks (presented in the paragraph ``Do Trackers Already Offer Any Advantage in FPV?'' of Section \ref{sec:resuluts} of the main paper.)

In the first experiment, we assessed the capabilities of continuous object localization of an object detector, as this method is usually exploited in many FPV pipelines. We executed the EK-55 trained Faster-R-CNN \cite{EK55} on all the frames of \datasetname, by recording in each frame the bounding box of the detection having the same class of the target and the highest confidence score.
Such bounding box predictions were then compared to the ground truth annotations \AF{using the considered tracking evaluation measures.} This experimental strategy respects the evaluation procedure of the OPE protocol. In this way, we can compare the performance of the tracking approach and the detection approach in providing localization and temporal reference of/to objects.

In the second experiment, we evaluated the impact of trackers in a video-based hand-object interaction detection setting. Since this paper is focused on objects (visual object tracking), we restricted our study in evaluating the detection of the objects involved in the interactions.
To this aim, we first built tracks of hand-object interactions over the sequences of \datasetname. The hand-object interaction detector Hands-in-contact \cite{Shan2020} has been executed to obtain sparse interaction detections that involved the object defined by the ground-truths of \datasetname. 
Clusters of detections have been then set to form separate tracks if the interval between two detections was longer than 30 frames. 
The missing detections within a cluster have been filled with the \datasetname's object ground-truth bounding boxes and the most frequent interaction state (i.e. if the object was in interaction with the left hand, the right hand, or both) appearing in the cluster. 
Once had these references, we ran the trackers in an OPE-like fashion. Each tracker was initialized in the first frame of a track with the object detection given by Hands-in-contact \cite{Shan2020}, and then let run for the other frames of the track. We then evaluated the performance in a track by the normalized count of frames having intersection-over-union $\geq 0.5$ with the object's ground-truth. The overall result is obtained by averaging the outcomes of all tracks. This experimental procedure gives us an estimate of the accuracy of the hand-object interaction detection system if trackers would have been included in its pipeline. More interestingly, it allows also to build a ranking of the trackers based on the results of a downstream application.

\begin{table}[t]
\fontsize{8}{9}\selectfont
	\centering
	\caption{Performance achieved by the 33 benchmarked trackers on \datasetname\ using the RTE protocol.}
	\label{tab:realtime32}
	\begin{tabular}{l | c c c c }
		\toprule
		
		Tracker & FPS & SS & NPS & GSR \\

		\midrule
		Ocean & 21 & 0.365 & 0.358 & 0.294 \\
		SiamBAN & 24 & 0.360 & 0.366 & 0.313 \\
		SiamRPN++ & 23 & 0.362 & 0.356 & 0.293 \\
		PrDiMP & 13 & 0.352 & 0.349 & 0.243 \\
		DiMP & 16 & 0.336 & 0.331 & 0.224 \\
		SiamMask & 23 & 0.335 & 0.333 & 0.298 \\
		SiamFC++ & 45 & 0.330 & 0.331 & 0.308 \\
		SiamDW & 32 & 0.327 & 0.334 & 0.317 \\
		KYS & 12 & 0.327 & 0.317 & 0.237 \\
		ATOM & 15 & 0.319 & 0.312 & 0.179 \\
		UpdateNet & 21 & 0.311 & 0.297 & 0.295 \\
		DCFNet & 49 & 0.299 & 0.286 & 0.335 \\
		TRASFUST & 13 & 0.296 & 0.270 & 0.185 \\
		SiamFC & 34 & 0.293 & 0.295 & 0.280 \\
		LTMU & 8 & 0.284 & 0.257 & 0.169 \\
		D3S & 16 & 0.276 & 0.263 & 0.182 \\
		BACF & 9 & 0.276 & 0.262 & 0.234 \\
		SPLT & 8 & 0.265 & 0.247 & 0.203 \\
		STRCF & 10 & 0.264 & 0.250 & 0.218 \\
		DSLT & 7 & 0.260 & 0.234 & 0.211 \\
		ECO & 15 & 0.252 & 0.231 & 0.173 \\
		GlobalTrack & 8 & 0.253 & 0.227 & 0.139 \\
		MCCTH & 8 & 0.251 & 0.231 & 0.232 \\
		Staple & 13 & 0.249 & 0.236 & 0.169 \\
		GOTURN & 44 & 0.247 & 0.242 & 0.119 \\
		MOSSE & 26 & 0.227 & 0.190 & 0.141 \\
        LTMU-H & 4 & 0.213 & 0.174 & 0.161 \\
        MetaCrest & 8 & 0.207 & 0.175 & 0.165 \\
        LTMU-F & 4 & 0.205 & 0.161 & 0.162 \\
        VITAL & 4 & 0.204 & 0.165 & 0.158 \\
        DSST & 2 & 0.191 & 0.145 & 0.161 \\
        KCF & 6 & 0.186 & 0.157 & 0.177 \\
        MDNet & 1 & 0.185 & 0.140 & 0.161 \\

		\bottomrule		
\end{tabular}
\end{table}

\section{Additional Results}
\label{sec:addres}

\pgraph{MSE Protocol Results.}
Figure \ref{fig:resultsmse} reports the overall performance of the 33 benchmarked trackers on \datasetname\ using the MSE protocol. The overall low performances of all the trackers confirm the conclusions achieved using the OPE protocol. The FPV setting introduces challenging factors for current visual trackers.

\begin{figure*}[t]%
\centering
\includegraphics[width=\linewidth]{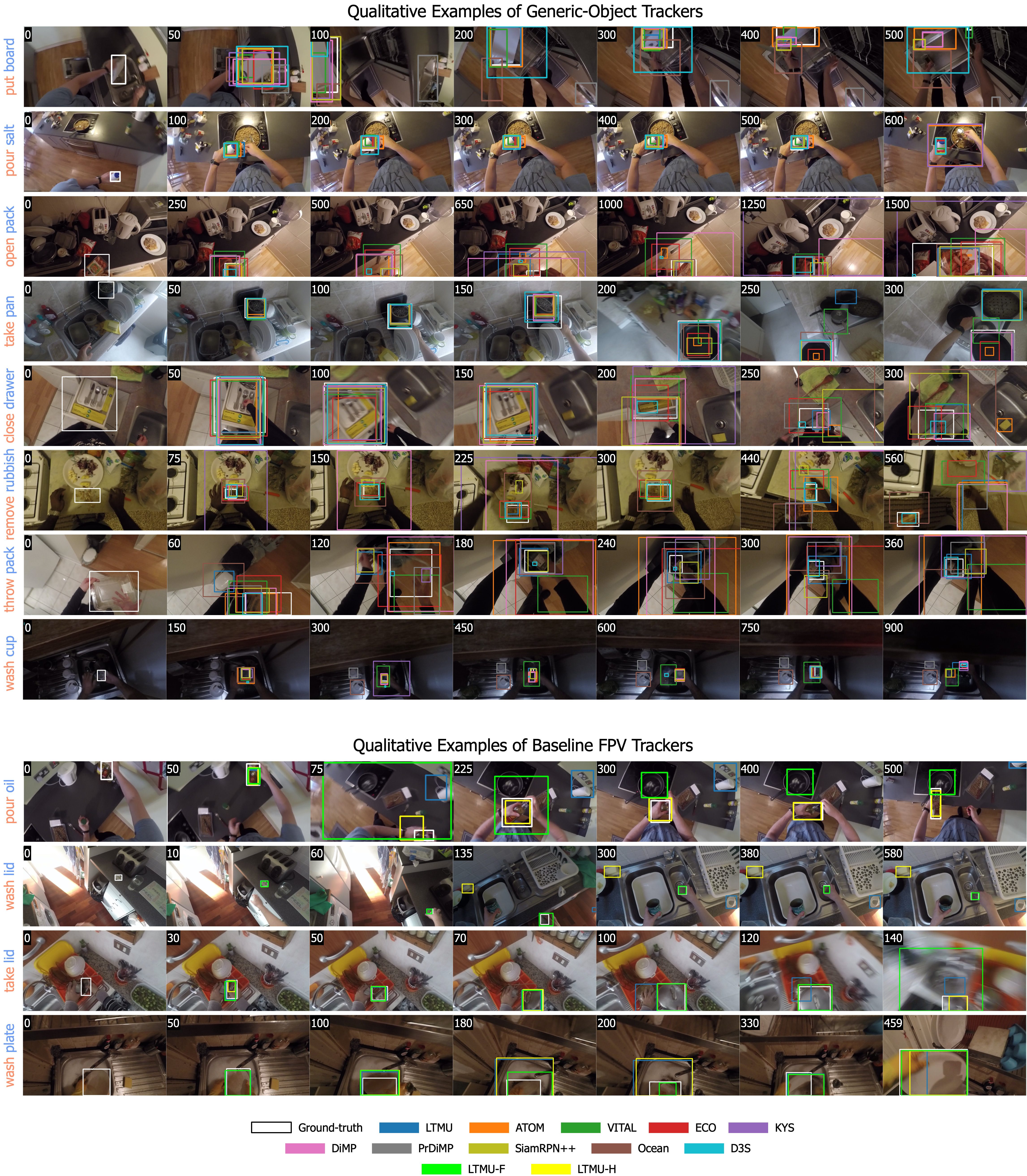}
\caption{Qualitative results of some of the studied trackers on the proposed \datasetname\ dataset. The first 7 rows of images show the qualitative performance of 10 of the selected generic-object trackers, while the last 4 rows show the results of the baseline FPV trackers LTMU-F and LTMU-H in comparison with LTMU. 
For a better visualization, a video can be found at \videolink.
}
\label{fig:qualitative}
\end{figure*}

\pgraph{Qualitative Examples}
The first 7 rows of images of Figure \ref{fig:qualitative} present qualitative results of 10 of the generic-object trackers in comparison with the ground-truth (which is identified by the white rectangles). The action performed by the camera wearer is also reported for each sequence. The remaining 4 rows show the qualitative performance of the FPV baseline trackers LTMU-F and LTMU-H in comparison with LTMU and the ground-truth.
For a better visualization, a video can be found at \videolink.

\begin{figure*}[t]%
\centering
\includegraphics[width=.8\linewidth]{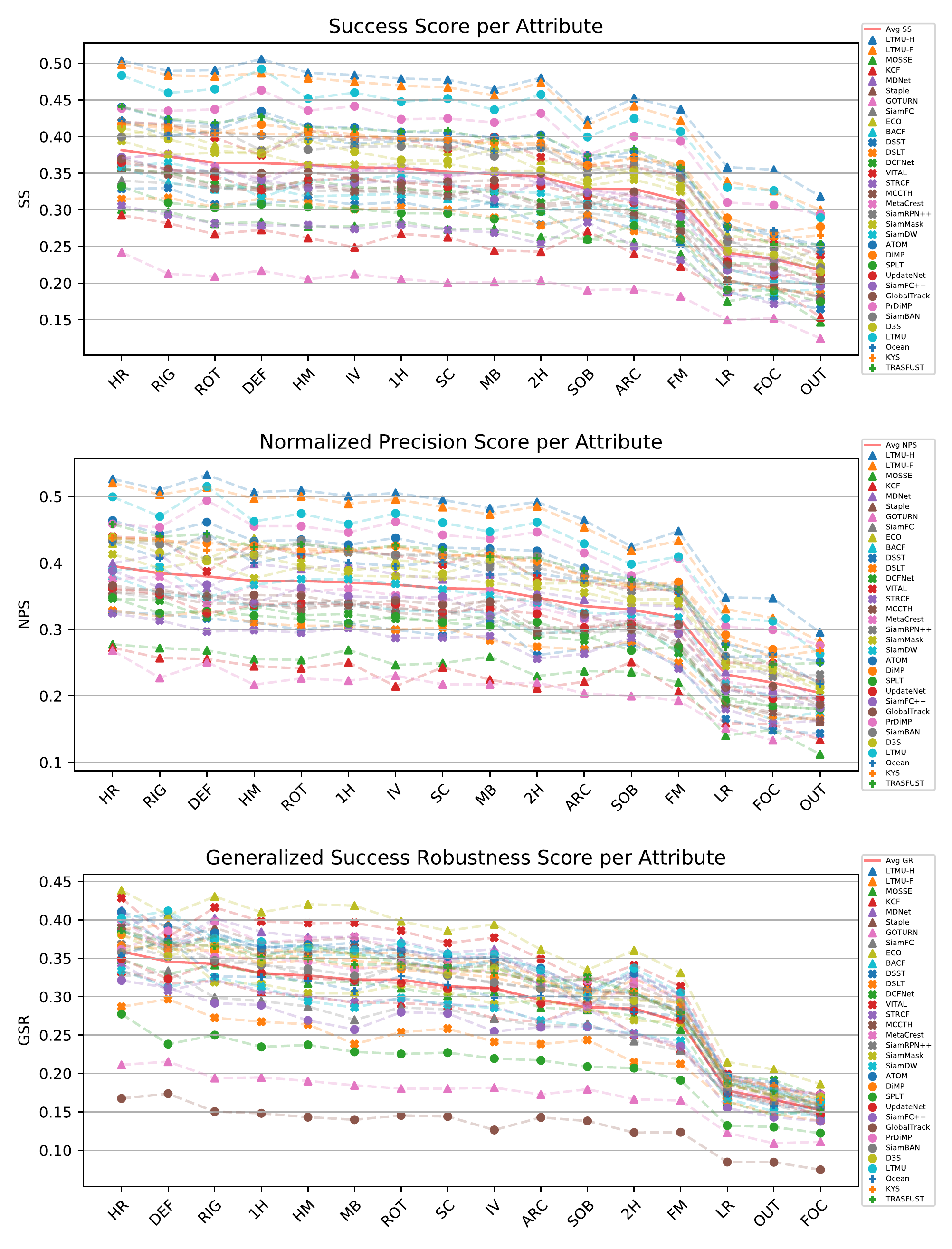}
\caption{SS, NPS, and GSR results per sequence attribute achieved by the 33 benchmarks on the \datasetname\ benchmark.}
\label{fig:resultsmse32attr}
\end{figure*}

\begin{figure*}[t]%
\centering
\includegraphics[width=.8\linewidth]{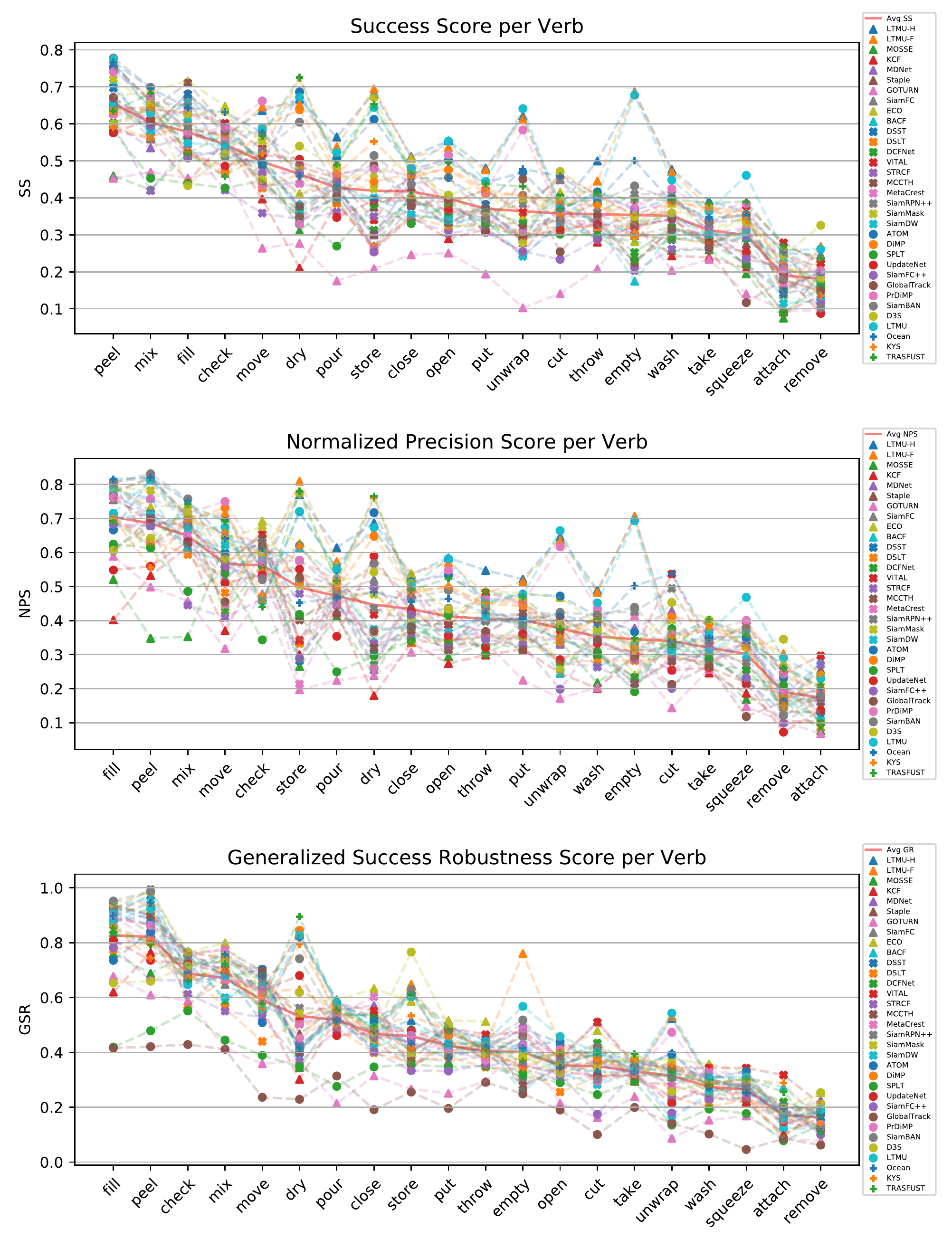}
\caption{SS, NPS, and GSR results achieved by the 33 benchmarks on the \datasetname\ benchmark considering each verb associated to the action performed by the camera wearer.}
\label{fig:resultsmse32verbs}
\end{figure*}

\begin{figure*}[t]%
\centering
\includegraphics[width=.8\linewidth]{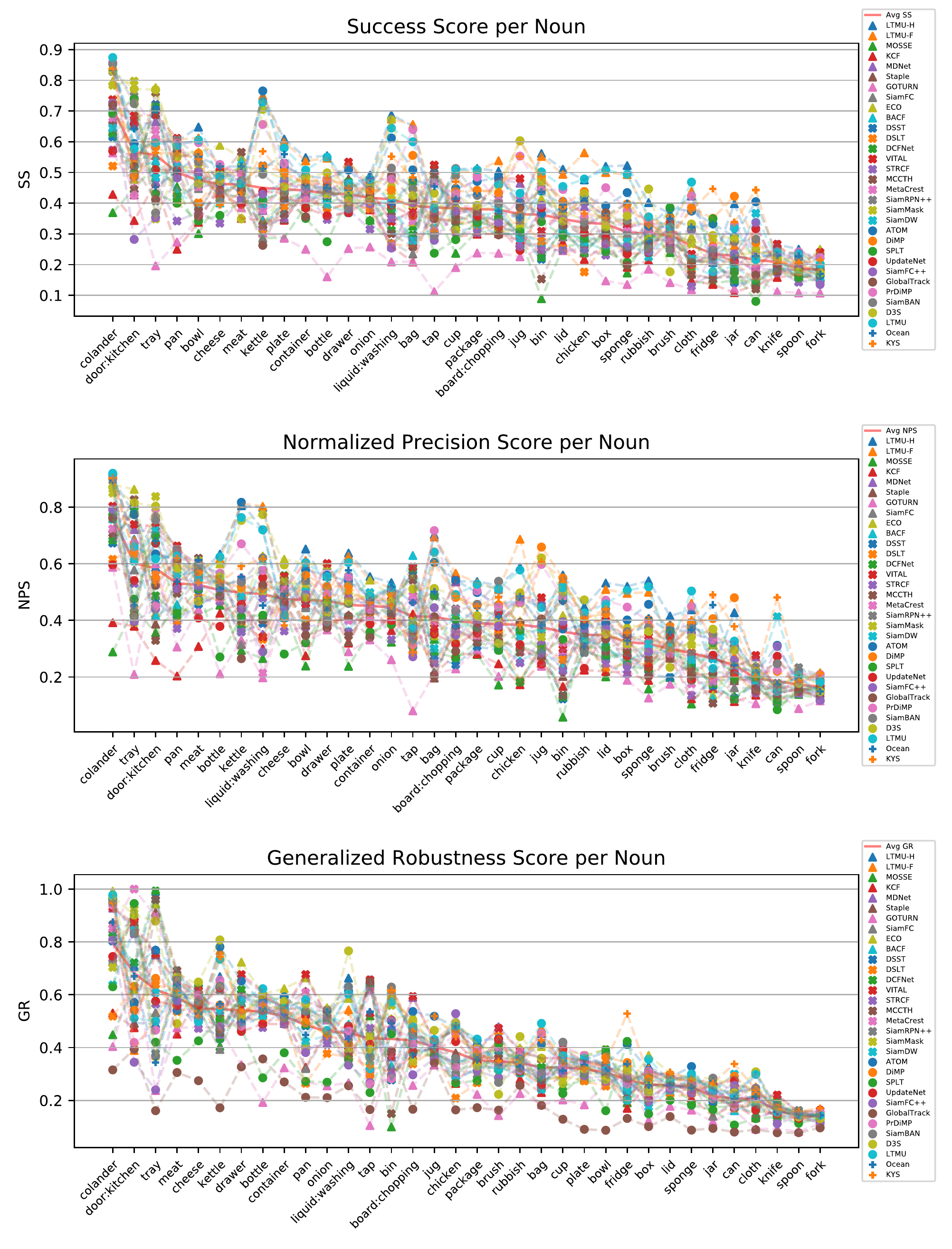}
\caption{SS, NPS, and GSR results achieved by the 33 benchmarks on the \datasetname\ benchmark considering the different target categories.}
\label{fig:resultsmse32nouns}
\end{figure*}

\pgraph{Per Attribute/Action Results.}
Figure \ref{fig:resultsmse32attr} presents the SS, NPS, and GSR scores achieved by the 33 trackers considering the attributes assigned to sequences. Similarly, Figures \ref{fig:resultsmse32verbs} and \ref{fig:resultsmse32nouns} report the results for the whole batch of trackers with respect to action verbs and target nouns.

\pgraph{RTE Protocol Results.}
Table \ref{tab:realtime32} reports the FPS, SS, NPS, and GSR performance of all 33 benchmarked trackers obtained using the RTE protocol. As stated in the main paper, offline siamese trackers emerge as the best solution in this scenario. Online deep discriminative trackers achieve comparable results  in SS and NPS, but demonstrate a larger drop in performance in the GSR score, showing that online learning mechanisms influence this performance in the real-time setting.

\begin{table}[t]
\fontsize{7}{8}\selectfont
	\centering
	\caption{Performance of the offline trackers SiamFC and SiamRPN++ on a subset of 50 sequences of \datasetname\ without and with fine-tuning on the remaining 100 videos.}
	\label{tab:finetuning}
	\setlength\tabcolsep{.15cm}
	\begin{tabular}{l | c | c  c  c | c  c  c  }
		\toprule
		\multirow{2}{*}{Tracker} & \multirow{2}{*}{Fine-tuning} & \multicolumn{3}{c|}{OPE}  & \multicolumn{3}{c}{MSE}\\
                    & & SS & NPS & GSR  & SS & NPS  & GSR \\
		\midrule
		\multirow{2}{*}{SiamFC}& \xmark & 0.311 & 0.332 & 0.317 & 0.307 & 0.317 & 0.307 \\
		& \checkmark & 0.267 & 0.275 & 0.278 & 0.287 & 0.305 & 0.292 \\
		
		\midrule
		
		\multirow{2}{*}{SiamRPN++}& \xmark & 0.384 & 0.395 & 0.377 & 0.367 & 0.385 & 0.333 \\
		& \checkmark & 0.348 & 0.407 & 0.313 & 0.336 & 0.406 & 0.314 \\
		\bottomrule		
\end{tabular}
\end{table}

\paragraph{Adaptation Of Offline Trackers.}
Many current visual trackers employ deep learning architectures. Among these, trackers based on siamese neural networks emerged as the most popular approaches nowadays. These trackers are said to be offline (e.g. SiamFC \cite{SiamFC}, SiamRPN++ \cite{SiamRPNpp}, SiamMask \cite{SiamMask}, SiamBAN \cite{SiamBAN}) because they are trained to track objects on large-scale tracking dataset \cite{ImageNet,TrackingNet,GOT10k,LaSOT}, and do not use online adaptation mechanisms at test time. In our evaluation, such trackers have been employed as they are described and trained in their original paper. Given their generally low performance, one could wonder how these trackers perform if knowledge about the FPV domain is exploited for learning.
Our \datasetname\ dataset, which is designed to evaluate the progress of visual tracking solutions in FPV, does not provide a large-scale database of learning examples as needed by these methods. Instead, it well aligns with real-world datasets where millions of frames are not available for training. In such scenarios, the reasonable options the machine learning community suggest are to use trackers as they are because of their general knowledge, or to adapt them through fine-tuning using a smaller training set. We tried the second strategy by randomly splitting \datasetname\ in a training and test set of 100 and 50 videos respectively. We fine-tuned the popular offline trackers SiamFC and SiamRPN++ on the training set according to their original learning strategy. We then tested the fine-tuned versions on the test set and the results are reported in Table \ref{tab:finetuning}. It shows that fine-tuning leads to substantial overfitting that cause the performance to drop in general. These outcomes prove the decision to evaluate offline trackers as they are is the right one given the current lack of large-scale FPV tracking datasets. Moreover, given the overall results presented in this paper, we hypothesize that visual tracking in FPV will require more than just large-scale training. We hope the results  presented in this section will encourage the community to work on domain adaptation techniques for offline trackers that are currently starting to be investigated \cite{Dunnhofer2021ral}.

\end{document}